\documentclass[final,12pt]{clear2024} 

\usepackage{amsmath,amsfonts,bm}

\def\eqref#1{equation~\ref{#1}}

\def\1{\bm{1}}

\def\vc{{\bm{c}}}

\def\vu{{\bm{u}}}

\def\vx{{\bm{x}}}

\DeclareMathAlphabet{\mathsfit}{\encodingdefault}{\sfdefault}{m}{sl}
\SetMathAlphabet{\mathsfit}{bold}{\encodingdefault}{\sfdefault}{bx}{n}

\def\sR{{\mathbb{R}}}

\def\sX{{\mathbb{X}}}
\def\sY{{\mathbb{Y}}}

\title[DiConStruct]{DiConStruct:\\Causal Concept-based Explanations through Black-Box Distillation}
\usepackage{times}
\usepackage{multirow}
\usepackage{booktabs}
\usepackage{makecell}
\usepackage{verbatim}
\usepackage{float}

\clearauthor{%
 \Name{Ricardo Moreira}\footnotemark[1] \Email{ricardo.moreira@feedzai.com} \\
 \Name{Jacopo Bono}\footnotemark[1] \Email{jacopo.bono@feedzai.com}\\
 \Name{Mário Cardoso}\footnotemark[1] \Email{mario.cardoso@feedzai.com}\\
 \Name{Pedro Saleiro}\footnotemark[1] \Email{pedro.saleiro@feedzai.com}\\
 \Name{Mário Figueiredo}\footnotemark[2] \Email{mario.figueiredo@tecnico.ulisboa.pt }\\
 \Name{Pedro Bizarro}\footnotemark[1] \Email{pedro.bizarro@feedzai.com}\\
 \addr {\footnotemark[1]Feedzai\\
 \footnotemark[2]Instituto de Telecomunicações, ELLIS Unit Lisbon, Instituto Superior Técnico, ULisboa, Portugal}%
}

\begin{document}

\maketitle

\begin{abstract}
Model interpretability plays a central role in human-AI decision-making systems. Ideally, explanations should be expressed using human-interpretable semantic concepts. Moreover, the causal relations between these concepts should be captured by the explainer to allow for reasoning about the explanations. Lastly, explanation methods should be efficient and not compromise the predictive task performance. Despite the recent rapid advances in AI explainability, as far as we know, no method yet fulfills these three desiderata. Indeed, mainstream methods for local concept explainability do not yield causal explanations and incur a trade-off between explainability and prediction accuracy. We present DiConStruct, an explanation method that is both concept-based and causal, which produces more interpretable local explanations in the form of structural causal models and concept attributions. Our explainer works as a distillation model to any black-box machine learning model by approximating its predictions while producing the respective explanations. Consequently, DiConStruct generates explanations efficiently while not impacting the black-box prediction task. We validate our method on an image dataset and a tabular dataset, showing that DiConStruct approximates the black-box models with higher fidelity than other concept explainability baselines, while providing explanations that include the causal relations between the concepts.
\end{abstract}

\begin{figure}[H]
   \subfigure[]{
        \includegraphics[width=0.6\textwidth]{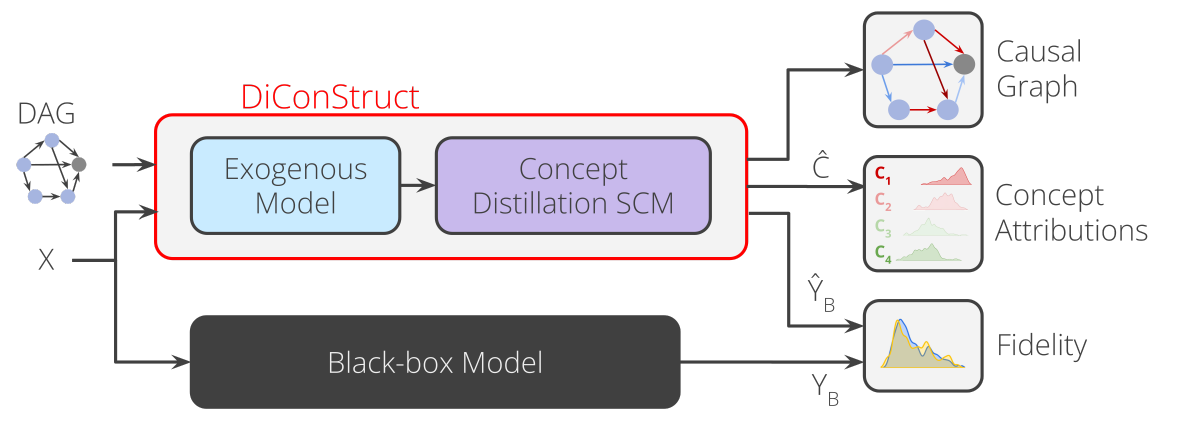}
        \label{fig:teaser_overview}
        }
    \hfill
   \subfigure[]{
        \includegraphics[width=0.4\textwidth]{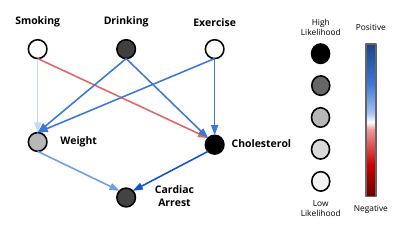}
        \label{fig:teaser_explanation}
        }
    \caption{\textbf{DiConStruct is a surrogate explainer for obtaining explanations in the form of (1) causal graphs relating concepts to black-box scores and (2) concept attributions.} (a) DiConStruct overview. (b) Toy example of a learned causal graph explanation for predicting cardiac arrest. For example, a low likelihood of smoking (white node) has negative impact on cholesterol (red edge) but positive impact on weight (blue edge).}
     \label{fig:teaser}
\end{figure}

\section{Introduction}
A considerable amount of work in explainable AI (XAI) is based on feature attribution \citep{burkart2021survey}. Those methods generate explanations for given machine learning (ML) model predictions as the set of contributions of each of the model inputs (features). In practice, feature-based explanations are often difficult to interpret \citep{kim2018interpretability} due to the absence of knowledge about the features' semantics and their connection to higher level concepts. This lack of interpretability makes feature attributions unsuitable for tasks where human-AI collaboration is vital, such as real-time human validation or correction of model decisions. Moreover, without incorporating the causal principles of interventions and counterfactuals, purely statistics-based ML systems are incapable of reasoning about explanations \citep{pearl2018theoretical}. The understanding of such limitations has motivated a considerable amount of work applying causality to XAI.
Although promising, most of these works still focus on feature-level explainability, such as feature attributions \citep{chattopadhyay2019neural, frye2020asymmetric, wang2021shapley, heskes2020causal, janzing2020feature, DBLP:journals/corr/abs-2008-00357, zhao2021causal, jung2022measuring}, feature-based counterfactual explanations \citep{karimi2020algorithmic, karimi2021algorithmic, bertossi2020asp}, and causally induced model structure \citep{geiger2022inducing}.

To improve the interpretability of explanations, there has been an increasing effort to develop methods that produce explanations understandable by non-AI experts. Towards this effort, leveraging higher-level concepts and producing concept-based explanations have become a popular method \citep{kim2018interpretability}. These concepts are human-defined meaningful characteristics that should describe well the most relevant patterns in the data. For example, in a dataset where the task is to classify birds, the wing color could be chosen as a concept. In this way, global or local explanations in the form of concept attributions are produced. Self-explaining methods \citep{koh2020concept, DBLP:conf/nips/Alvarez-MelisJ18} jointly learn the prediction and the explanation tasks with a single model, but have a potential disadvantage of creating a trade-off between the two tasks  \citep{DBLP:conf/iclr/BahadoriH21}. Such trade-off is especially undesirable in contexts where keeping the performance of the main prediction task is critical. Finally, apart from work by \citet{DBLP:journals/corr/abs-2006-02482}, which is limited to producing global explanations, concept explanation methods are still not incorporating causal principles, limiting their interpretability by ignoring the underlying causal relationships.

The problems mentioned above motivate us to develop an explanation method that incorporates the principles of causality and provides concept-based explanations, as depicted in Figure \ref{fig:teaser}. We aim to improve the quality of explanations produced by understanding the causal relationships between the concepts and predictions, in the form of a learned causal graph. Furthermore, we aim at creating a surrogate explainer instead of a self-explaining method, ensuring that the main prediction task is never compromised. In the following section, we provide further arguments for why we believe XAI research should be focusing on bringing together concept explanations and causal modeling. 

In summary, the main contribution of this work is a novel explainer, with the following key characteristics:
\begin{itemize}
    \item it is concept-based and causal, yielding more interpretable local explanations in the form of structural causal models (SCM) and concept attributions; 
    \item it is a surrogate model, hence not affecting the predictive performance of the ML model, while simultaneously producing explanations efficiently in a single forward pass.
\end{itemize}
Moreover, we validate our explainer in two different domains (image data and tabular data), showing that it approximates the black-box models with higher fidelity than other concept explainability baselines, while providing explanations that include the causal relations between the concepts.

The code of our experiments is publicly available \footnote{https://anonymous.4open.science/r/causal-diconstruct-4F18/}.

\section{DiConStruct}
\begin{figure}[t]
    \begin{center}
        \includegraphics[width=1.0\linewidth, keepaspectratio]{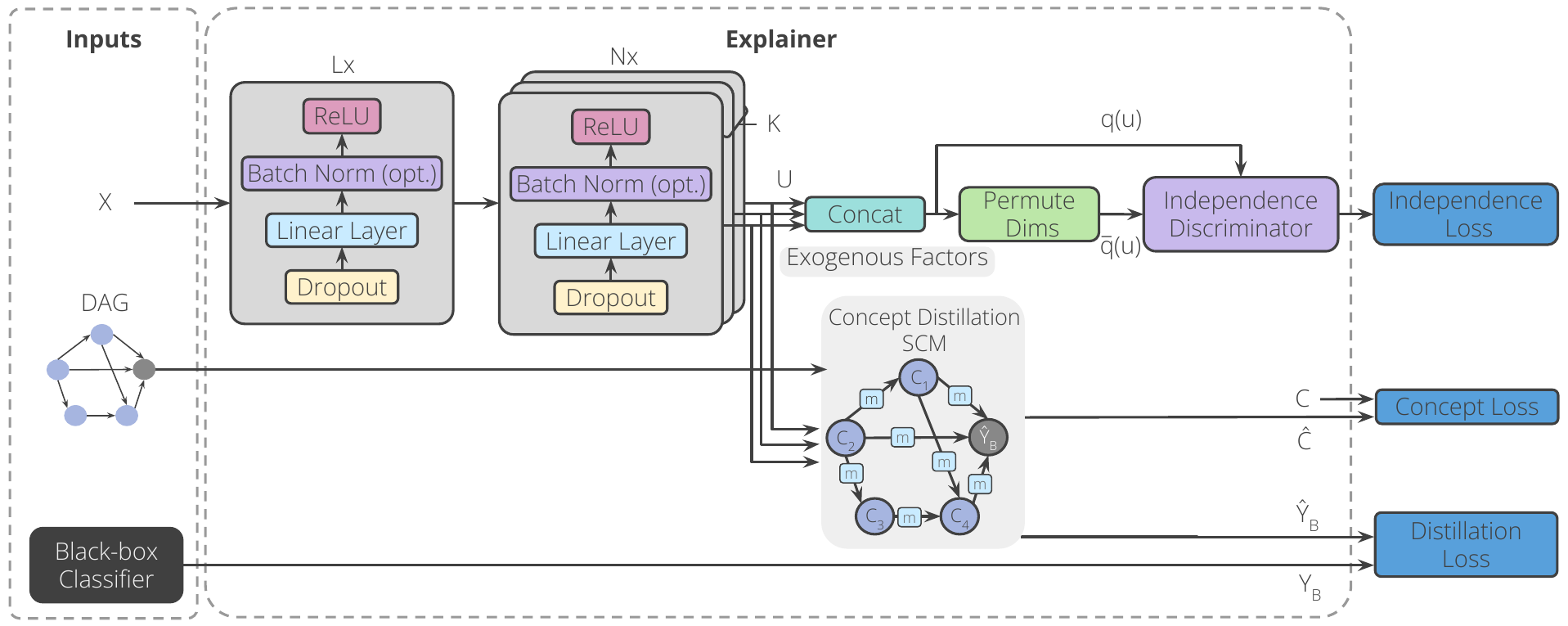}
    \end{center}
    \caption{DiConStruct architecture. On the left side are the inputs to our explainer: (1) an instance to explain $\vx$; (2) a DAG relating the concepts (Section \ref{sec:obtaining-a-causal-graph} discusses approaches to obtain this DAG); (3) a black-box classifier.}
    \label{fig:causal_diconstruct}
\end{figure}
\subsection{Preliminaries}
\label{sec:perliminaries}

In the same vein of what was proposed by \citet{DBLP:journals/corr/abs-2006-02482}, we model the causal mechanism $C \rightarrow \hat{Y}$, where $C$ are concepts and $\hat{Y}$ the black-box model predictions (see Appendix \ref{appendix:causal-view-explainability}). Our goal is to gain insights into what information the ML model learned during training (global explanations) as well as explaining the individual decision processes (local explanations). We frame our explainability problem as a specific knowledge distillation task, where the explainer acts as a surrogate model learning to mimic a black-box binary classifier $f_B$, trained to produce the score of a binary target, $f_B(\vx) = P(y=1 | \vx) = y_B$. The task of our explainer is, given some instance $\vx$, to produce high-fidelity estimates of $y_B$, while also correctly predicting the $K$ semantic concepts, $\{c_k\}_{k=1}^K$, associated with that instance. All these predictions are produced in a structural causal model (SCM) that models the joint distribution $P(C, Y_B)$. For each instance, the explainer produces a local SCM through corresponding local concept attributions. 

We define this explainer more formally as $\Phi : \sX \rightarrow \sY_B \times \sY_C$, where $\sX$ is the space of the model features, $\sY_B = [0, 1]$ is the space of the classifier estimates, $y_B$, corresponding to the distillation targets, and $\sY_C = \{0, 1\}^{K}$ is the space of binary labels for the $K$ semantic concepts. Function $\Phi$ is learned on both objectives mentioned in the previous paragraph using a dataset $\mathcal{D}_\phi = \bigl\{\bigl(\vx^{(i)}, y_B^{(i)}, \{c_k^{(i)}\}_{k=1}^{K}\bigr)\bigr\}_{i=1}^{M}$ with $M$ instances and a pre-defined directed acyclic graph (DAG): $\mathcal{G}_{C,Y_B}$. The DAG is assumed to be an input to our explainer (see Section \ref{sec:obtaining-a-causal-graph}). The explainer then produces estimates $\Phi(\vx) = (\hat{y}_B, \{\hat{c}_k\}_{k=1}^{K})$ and explanations in the form of an SCM. We refer to Appendix \ref{causality-definitions} for a formal definition of an SCM and its relation with a DAG.
Finally, one can obtain instance-wise concept attributions by performing counterfactual analysis on this causal graph (details in Section \ref{sec:concept-attributions}). 

We define 2 necessary conditions for the obtained explanations to be causally valid.\\
\textbf{$\bullet$ Exogenous independence}: the external variables of the SCM are so-called "exogenous" variables (definition in Appendix \ref{causality-definitions}), which are learned for each instance. For interpretability purposes, the exogenous variables should be mutually independent. If this condition is not satisfied, it implies that some other unmeasured variable is causing their dependence, meaning that not all relationships between the concept variables are being represented in the explainer's structural equations. This would make interventions and counterfactual analysis unfeasible, since not all required information would be available in the SCM. In other words, we require \emph{causal sufficiency} \citep{pearl2009causality} in order to perform counterfactual analysis. \\
\textbf{$\bullet$ Concept completeness}: for our explainer to approximate the black-box model, we require the causal parents of $Y_B$ in $\mathcal{G}_{C,Y_B}$ to fully explain the black-box behavior. For this to be possible, we need (1) the pre-defined concept variables to verify the \textit{completeness} criteria \citep{yeh2020completeness} and that (2) all causal mechanisms that are responsible for correctly approximating $Y_B$ to be in place in the DAG. Because our explainer is a surrogate model, the targets are not $Y$, but the predictions of a black-box model $Y_B$. In this scenario, the \textit{completeness score} proposed by \citet{yeh2020completeness} reduces to the accuracy of the concept model $g(\vc) = \hat{Y}_B$ (i.e. the model predicting
the label just given the concept scores) at predicting the targets $Y_B$\footnote{The term in the denominator of the original \textit{completeness score} definition will always be 1 (maximum) since we can always find a model that approximates $f_B(x)$ perfectly, like the black-box itself.}.

We name our explainer DiConStruct (from \emph{distillation}, \emph{concept-based} and \emph{structural} causal model); and it can be divided into two components (see Figure \ref{fig:causal_diconstruct}):\\
\textbf{$\bullet$ Exogenous model}: learns to predict independent exogenous variables $\vu = (u_1 \dots u_K)$ from $X$; \\
\textbf{$\bullet$ Concept distillation SCM}: learns the joint distribution $P(C, Y_B)$ in an SCM with partial structural assignments $m_{j,k}$ for each edge $(j,k)$.

In the following subsections, we define our exogenous model, propose how the causal DAG $\mathcal{G}_{C,Y_B}$ can be obtained for modelling $P(C, Y_B)$, define our concept distillation SCM, and, finally, show how we obtain \textit{concept attributions}.

\subsection{Exogenous Model with an independence objective}
\label{section:exogenous-model}
To guarantee the exogenous independence condition (Section \ref{sec:perliminaries}), we learn our exogenous model, defined as $f_E : \sX \rightarrow \sY_C$, imposing an independence objective on its outputs. 
The exogenous model is designed as a multi-output feedforward neural network consisting of $L$ common layer blocks and $N$ concept-specific layer blocks (see Figure \ref{fig:causal_diconstruct}). Each layer block is composed of a dropout layer, a linear layer, an optional batch normalization layer \citep{ioffe2015batch}, and a ReLU activation function. The outputs are subject to an independence objective, for which we followed prior art by \citet{kim2018disentangling}. Essentially, we want to make the joint distribution of exogenous variables, $q(\vu) = p(u_1, ..., u_K)$, indistinguishable from $\Bar{q}(\vu) = \prod_{k=1}^K p(u_k)$. As proposed by \citet{arcones1992bootstrap}, we sample from $\Bar{q}(\vu)$ by taking samples $\vu \sim q(\vu)$ and randomly shuffling the components of $\vu$ (i.e. exogenous variables) across a batch of instances. For each component, a different seed is used for shuffling, breaking all existing dependencies in $q(\vu)$. We use an independence discriminator model $f_I : \sY_C \rightarrow [0, 1]$, trained to distinguish $q(\vu)$ from $\Bar{q}(\vu)$, as a means to approximate the Kullback–Leibler divergence between $q(\vu)$ and $\Bar{q}(\vu)$ by \citep{arcones1992bootstrap,nguyen2010estimating}:
\begin{align}
\begin{split}
    D_{KL}(q(\vu)||\Bar{q}(\vu)) = \mathbb{E}_{q(\vu)} \biggl[ \log\frac{q(\vu)}{\Bar{q}(\vu)} \biggr] \approx \mathbb{E}_{q(\vu)} \biggl[\log{\frac{f_I(\vu)}{1 - f_I(\vu)}}\biggr] := \mathcal{L}_E.
\end{split} \label{eq:loss_e}
\end{align}

\subsection{Obtaining a Causal DAG for $P(C,Y_B)$}
\label{sec:obtaining-a-causal-graph}

As mentioned previously, to achieve high-fidelity approximations of the black-box score, we require a close to \textit{complete} concept definition. The task of defining which concepts are employed for explaining the black-box model is commonly performed by domain experts. Given the concepts, we then propose 2 different ways of obtaining $\mathcal{G}_{C,Y_B}$: (1) from expert knowledge and (2) using causal discovery. While option (2) is more attainable since it depends solely on data, we stress that it still requires concept definition to be made beforehand. Furthermore, causal discovery will not correct for any possible lack of \textit{completeness} of the concept definition, as it only performs structure learning, not discovery of new concepts. It is important to highlight that the concepts and causal DAG are given as inputs to our method; their discovery is not part of our method. 

\subsection{Global and Local Structural Causal Concept Distillation}
Given the exogenous variables (Section \ref{section:exogenous-model}) and a causal graph (Section \ref{sec:obtaining-a-causal-graph}), we now define our second component of the DiConStruct explainer: the concept distillation SCM. This model takes any causal graph $\mathcal{G}_{C, Y_B}$ and creates a set of partial structural assignments $m_{j,k}$ for each graph edge $(j, k)$. These partial structural assignments encode the strengths of the direct effects of the causal mechanisms $C_j \rightarrow C_k$ and $C_j \rightarrow Y_B$. We model those causal mechanisms in two ways. Firstly, in a global way, by assuming a linear dependency on \textit{logits} of the different variables in the graph,
\begin{equation}
    m_{j,k}(\hat{c}_j) = \sigma^{-1}(\hat{c}_j)w_{j,k},
\end{equation}
where $k=1,...,K$, $\sigma$ is the sigmoid function, $\hat{c}_j$ is the structural assignment of concept $j$ and will be defined below in equation \ref{eq:struct_assignment}, and $w_{j,k}$ can be thought of as global edge weights. Secondly, in a local way, by introducing the non-linear weighting functions $W_{j,k}: [0,1]^K \times \sY_B \rightarrow \sR$, which receive as inputs the set of exogenous variables $U$, and the true black-box score,
\begin{equation}
    m_{j,k}(\hat{c}_j, \vu, y_B) = \sigma^{-1}(\hat{c}_j) W_{j,k}(\vu, y_B),
\end{equation} 
where the $W_{j,k}(\vu, y_B)$ define the edge weights for each instance.
The black-box model score $y_B$ is included as input to the weighting functions as a means to improve the distillation fidelity. We create the partial structural assignments following the same formulation both for edges of type $C_j \rightarrow C_k$ and type $C_j \rightarrow Y_B$. Finally, we obtain the structural assignments for each concept in the graph as
\begin{equation}
    \hat{c}_k := f_k(\mathbf{PA}_k,u_k) = \sigma\biggl(b_k + \sigma^{-1}(u_k) + \sum_{j \in \mathbf{PA}_k} m_{j,k}\biggr), \label{eq:struct_assignment}
\end{equation}
where $\mathbf{PA}_k$ is the index set of causal parents of $C_k$ in $\mathcal{G}_{C, \hat{Y}_B}$. Importantly, these structural assignments are computed by traversing the DAG in topological order, starting from the nodes without parents. Depending on the instantiation of the explainer, the partial $m_{j,k}$ can take either the global or the local formulation. We use an analogous structural assignment for $\hat{Y}_B$, without the logit term since no exogenous variable is learned for $Y_B$.

The structural assignments presented above depend on static weights $w_{j,k}$ (or $w_{j,B}$) or weighting functions $W_{j,k}$ (or $W_{j,B}$) and biases $b_k$ (or $b_B$). All these components are learned during training and fixed at inference time. To learn these components, we use two binary cross-entropy (BCE) losses between the predictions $\hat{c}_k$ and $\hat{y}_B$ and the targets $c_k$ and $y_B$. We name the losses $\mathcal{L}_C$ (explainability loss) and $\mathcal{L}_D$ (distillation loss), respectively:

\begin{align}
    & \mathcal{L}_C = \sum_{i=1}^{M} \sum_{k=1}^{K} \sum_{c_k} c_k^{(i)} \log_2(\hat{c}_k^{(i)}), & \hspace{10pt} \text{(explainability)} \label{eq:loss_c} \\
    & \mathcal{L}_D = \sum_{i=1}^{M} \sum_{y_B} y_B^{(i)} \log_2(\hat{y}_B^{(i)}), & \hspace{10pt} \text{(distillation)} \label{eq:loss_d}
\end{align}
with $M$ denoting the instances in the evaluation set.

\subsection{Learning the DiConStruct explainer}
All components of the DiConStruct explainer are learned jointly through gradient descent (GD) and backpropagation using the combined losses from equations \ref{eq:loss_e}, \ref{eq:loss_c} and \ref{eq:loss_d}:
\begin{equation}
    \mathcal{L} = \gamma\mathcal{L}_E + \beta\mathcal{L}_C + \mathcal{L}_D,
\end{equation}
where $\beta$ and $\gamma$ are tunable hyperparameters. We point out that the gradients of $\mathcal{L}_E$ only affect the exogenous component. Because $\mathcal{L}_E$ consists of an expectation over a sample of instances, we compute it in batches, therefore restricting our method to learn through batch or mini-batch GD.

To train the independence discriminator $f_I$ to distinguish $q(\vu)$ from $\Bar{q}(\vu)$, we also use mini-batch GD with a BCE loss. The GD updates performed on the discriminator are interleaved with the updates of the other model components. Optionally, the updates of $f_I$ might occur at a configurable frequency, set by a parameter $\theta$.

\subsection{Concept Attributions from Counterfactual Statements}
\label{sec:concept-attributions}

For a given instance $\vx^{(i)}$, to measure the total causal effect of a given concept $C_k$ on $Y_B$, we can use the following counterfactual statements from our SCM $\mathfrak{C}^{(i)}$:
\begin{equation}
    P^{\mathfrak{C}^{(i)}}_{do(C_k := a)}(Y_B^{(i)} = 1), \label{eq:counterfactual}
\end{equation}
which, depending on the value of $a \in \{0, 1\}$, aims to answer different questions. If $a = 1$, the counterfactual question is: \textit{what would have been the probability of a positive prediction of the black-box, had the concept $C_k$ been present in the instance $\vx^{(i)}$}. Conversely, if $a = 0$ the counterfactual question becomes: \textit{what would have been the probability of a positive prediction of the black-box, had the concept $C_k$ \textbf{not} been present in the instance $\vx^{(i)}$}. These two statements are computed following the three steps for computing counterfactuals suggested by \citet{pearl2009causality}:\\
$\bullet $ \textbf{Abduction}: we use the predicted $\vu^{(i)}$ as the necessary exogenous variables for this instance;\\
$\qquad \bullet $ \textbf{Action}: for the assignment $do(C_k := 1/0)$, we directly modify  $\mathfrak{C}^{(i)}$ by removing all causal parents from $C_k$ and setting its value to either $0$ or $1$;\\
$\qquad \bullet $ \textbf{Prediction}: leaving all the remaining concept predictions fixed, we compute the new value of $\hat{y}_B$, which corresponds to the counterfactual quantity. 

We then obtain the concept attributions as total concept effects (TCE):
\begin{equation}
\label{eq:tce}
    \mbox{TCE}^{(i)}(C_k := a) = P_{do(C_k := a)}(Y_B^{(i)} = 1) - P(Y_B^{(i)} = 1), \hspace{10.0pt} a \in \{0, 1\},
\end{equation}
where $P(Y_B^{(i)} = 1)$ corresponds to the \textit{factual} probability of the black-box classifier prediction to be positive for the instance $\vx^{(i)}$.

\section{Experimental Setup}

\subsection{Datasets and black-box models}
\label{datasets-and-blackboxes}

\textbf{CUB-200-2011.} We validate our method using the Caltech-UCSD Birds-200-2011 \citep{wah2011caltech} dataset, consisting of 11,788 bird images in 200 subcategories (with 316 annotated concepts). We split this into train, validation, and test, with 4818, 1176, and 5794 images, respectively. The binary classification task is detecting the $warbler$ bird category, the most prevalent one (12.63\% prevalence). Additionally, since concepts have multiple values, we binarize the concepts by assigning 1 to the most prevalent attribute value and 0 otherwise (see Appendix \ref{sec:concept-labels}). To train a black-box model, we use a pre-trained ResNet-34 \citep{he2016deep} model, removing the last 8 layers and attaching a fully connected network with varying layer dimensions.

\textbf{Merchant Fraud.} We use a privately held online retailer fraud detection dataset, with approximately 5.5M transactions, 2\% of which represent fraudulent behavior. Concept labels are assigned using a weak supervision method starting from a small sample of human-labeled data (Appendix \ref{sec:concept-labels}).  Three sequential subsets form the train, validation and test sets, composed of 4.8M, 200k, and 1M instances, respectively. We trained 2 different black-box models on this dataset, on the task of predicting the binary outcome: \textit{fraud}/\textit{legit}. The first black-box model is a LightGBM \citep{ke2017lightgbm} and the second a feedforward neural network (NN). Because our goal is model explainability, we create a sample of the transactions that would most likely need to be explained, containing the transactions of highest uncertainty for the black-box fraud detection system. This resulted in 460k instances for training, 17k for validation, and 107k for testing. The fraud rate in this sample is 21.6\%. Our DiConStruct explainers and baselines were trained and tested using these sampled datasets.

\subsection{Evaluation Metrics}
\label{sec:evaluation-metrics}
\textbf{Main Task Performance.} Given the class imbalance, we chose to use the metric true positive rate (TPR) evaluated at a fixed false positive rate (FPR), which we set to be $5\%$. In other words, this translates to maximizing the amount of correctly classified minority class events (label positives) while incorrectly classifying at most $5\%$ of the majority class (label negatives).\\
\textbf{Fidelity.} We measure how close the prediction of our distillation model is to the black-box model prediction. We use the 1 - MAE (mean absolute error) to quantify the distillation quality.\\
\textbf{Concept Performance.} We measure this performance by calculating the prediction accuracy for each concept, corresponding to the error rate for binary decisions using 0.5 as probability threshold, and then averaging that over the $K$ concepts.

\subsection{Full Setup}
We distinguish four DiConStruct variants corresponding to the four combinations of two configurations: exogenous model \emph{with or without the independence} objective; \emph{local or global} concept distillation SCM configurations.

To obtain the causal DAG, we resort to causal discovery methods for each of the three black-box models: the PC algorithm \citep{spirtes2000causation}; the ICA-LiNGAM method \citep{shimizu2006linear}; the NO TEARS algorithm \citep{zheng2018dags}. We restricted the causal discovery to DAGs in which the black-box score variable has no children, since the score of the black-box model is never a cause for any of the concepts. Additionally, we added a \textit{trivial} DAG baseline in which all concepts are direct parents of $\hat{Y}_B$ without edges between any of the concepts.

Combining these four variants, with the four causal graphs obtained for each of the three black-box models (CUB NN, Merchant Fraud NN, Merchant Fraud LGBM), yields a total of 48 experiments. In each experiment, we perform a random hyperparameter search of 50 iterations, resulting in a total of 2400 trained DiConStruct explainers. Additional details regarding hyperparameters are provided in Appendix \ref{hyperparameter-details}. 

\section{Results}

\subsection{Baselines}
We compare our DiConStruct method with the following baselines:\\
\textbf{Single task – task performance:} the best black-box models on the main classification task.\\
\textbf{Single task – concept prediction:} we train a model with the same structure as the exogenous model (excluding the independence objective components). The outputs of the $K$ concept-specific layers are directly compared with the true labels using the concept loss $\mathcal{L}_C$. \\
\textbf{Single task – knowledge distillation:} for the knowledge distillation task, we train a standalone feedforward neural network with a BCE loss. \\
\textbf{Concept bottleneck models (CBMs):} we reproduced the concept bottleneck model framework \citep{koh2020concept}. The CBM is a self-explaining model given by $f(g(x))$, where $g: \sX \rightarrow \sY_C$ and $f: \sY_C \rightarrow \sY_B$. Following the \textit{joint} approach proposed by \citet{koh2020concept}, we set the $\lambda$ parameter to 1, giving equal weight to the main task and concept prediction losses. We varied the same parameters on $g$ and $f$ as we did for our DiConStruct explainers. Since DiConStruct is a post hoc explainer, we also trained a CBM as a surrogate explainer by passing the black-box scores as targets for $f$. We therefore have two types of CBM baselines: (1) the \emph{joint CBM} as an intrinsically explainable method; and (2) the \emph{distillation joint CBM} as a post hoc explainer.

For each combination of baseline, black-box model (if applicable), and dataset we perform a separate random hyperparameter search for 50 iterations for the single task baselines and 50 for the concept bottleneck baselines. Information about the hyperparameters is provided in Appendix \ref{hyperparameter-details}.

\subsection{Fidelity and concept accuracy performances}

\begin{figure}[h]
    \begin{center}
        \includegraphics[width=1.0\linewidth, keepaspectratio]{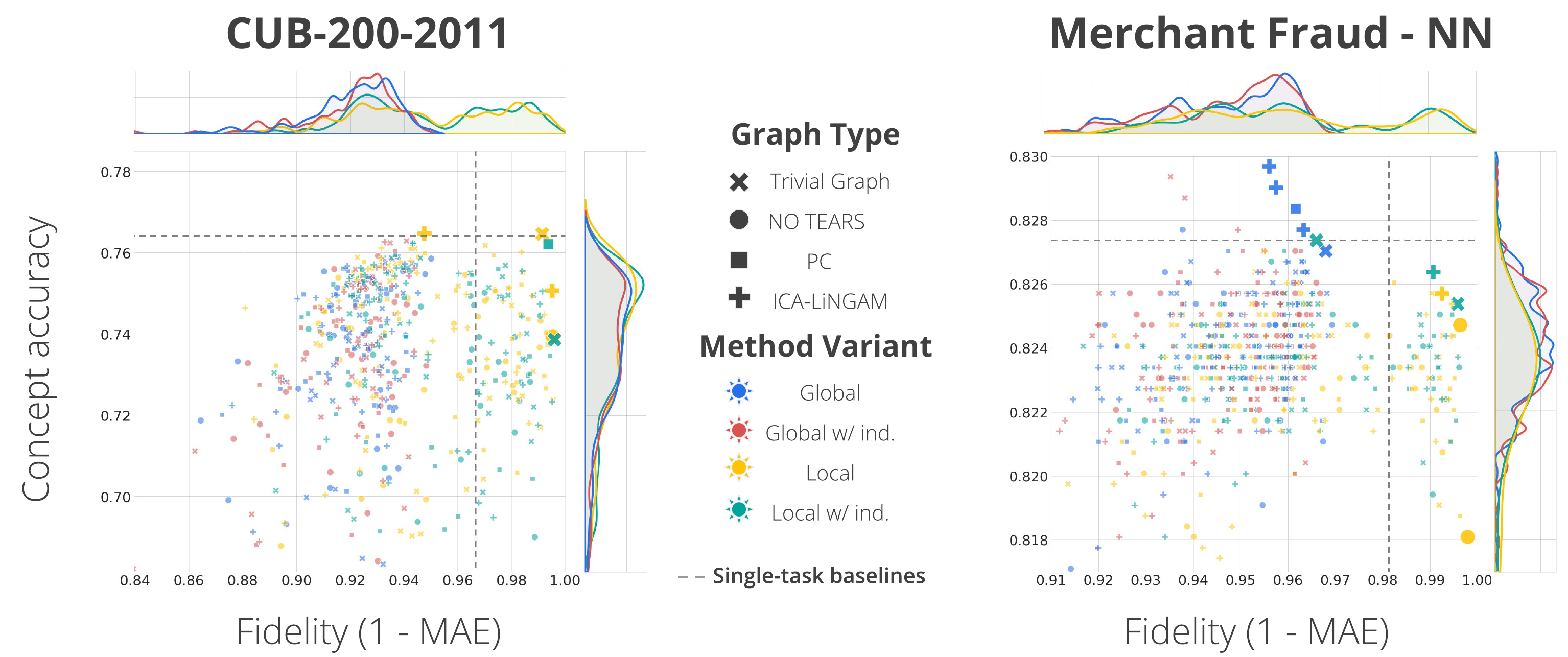}
    \end{center}
    \caption{Distribution of fidelity and concept accuracy for the CUB-200-2011 (left) and Merchant Fraud (right) black-boxes. Each plot shows the result of 16 experiments of 50 trials, and each marker is a trained DiConStruct explainer. Models on the Pareto efficiency frontier are highlighted with bigger markers.}
    \label{fig:performances}
\end{figure}

We estimate the variability of hyperparameter search by bootstrapping over the trained explainers. Each bootstrap trial consists of 20 random draws from the pool of 50 trained models per experiment. The best performing model (maximizing the sum of the validation fidelity and concept accuracy) of each trial is then separately selected. This process was repeated for 1000 trials to obtain a distribution of results for each metric. Table \ref{tab:best_dag_results_table} shows the results for the DAGs attaining the highest sum of mean fidelity and mean concept accuracy on each method variant. For comparison, we add the results of the CBMs and the single task baselines, for which we performed the same bootstrap analysis. In terms of concept accuracy, DiConStruct variants are comparable with both the CBMs and the single task baselines. Being a surrogate explainer, the main task performance of our DiConStruct variants is the same as the best black-box model. With the original CBMs, we observe a significantly lower task performance resulting from the tradeoff caused by the joint training of both tasks. This drop in performance can be mitigated by training the CBM as a surrogate explainer (distillation CBM). Comparing the global variants of our method with the baselines on the fidelity metric, we observe the performances are comparable. Local variants of DiConStruct, on the other hand, outperform the baselines in terms of fidelity for the two dataset and black-box models. As mentioned previously, this is due to the use of the black-box score as an input to the local concept distillation SCM. Similar conclusions can be drawn from Figure \ref{fig:performances}, where we see that some configurations outperform the single task baselines individually, particularly notable for the local variants. The full set of results considering each variant and DAG is provided in Appendix \ref{complete-performance-results}.

\setlength{\aboverulesep}{0pt}
\setlength{\belowrulesep}{0pt}

\begin{table}[h]
\centering
\resizebox*{\linewidth}{!}{
\begin{tabular}[t]{cll|ccc | ccc}
& & & \multicolumn{3}{|c}{
            \textbf{Validation}} & \multicolumn{3}{|c}{
            \textbf{Test}} \\
\cmidrule{2-9}
    &
    \multicolumn{1}{c}{\textbf{Model}} & 
    \textbf{Variant} & 
    \makecell{\textbf{Task} \\ \textbf{Perf. (\%)}} & 
    \makecell{\textbf{Concept} \\ \textbf{Perf. (\%)}} &   
    \textbf{Fidelity (\%)} & 
    \makecell{\textbf{Task} \\ \textbf{Perf. (\%)}} & 
    \makecell{\textbf{Concept} \\ \textbf{Perf. (\%)}} &  
    \textbf{Fidelity (\%)} \\
\toprule 
\multirow{8}{*}{\rotatebox[origin=c]{90}{\textbf{CUB-200-2011}}} & \multirow{4}{*}{Ours} & Global & \multirow{4}{*}{77.85} & 75.58 $\pm$ 0.65 & 93.52 $\pm$ 0.77 & \multirow{4}{*}{79.05} & 75.44 $\pm$ 0.44 & 94.3 $\pm$ 0.8 \\
& & Global w/ Ind. & & 75.61 $\pm$ 0.69 & 93.16 $\pm$ 0.75 & & 75.43 $\pm$ 0.65 & 93.83 $\pm$ 0.64  \\
& & Local & & 75.25 $\pm$ 0.76 & 98.72 $\pm$ 0.79 & & 75.11 $\pm$ 0.63 & 98.79 $\pm$ 0.74\\
& & Local w/ Ind. & & 75.05 $\pm$ 1.08 & 98.78 $\pm$ 0.86 & & 74.89 $\pm$ 1.19 & 98.83 $\pm$ 0.8 \\
\cmidrule{2-9}
& \multirow{4}{*}{Baselines} & Joint CBM ($\lambda$ = 1)  & 79.25 $\pm$ 0.98 & 75.57 $\pm$ 0.46 & - & 67.33 $\pm$ 2.13 & 75.76 $\pm$ 0.55 & -\\
& & Distill. Joint CBM ($\lambda$ = 1) & 77.85 & 75.48 $\pm$ 0.53 & 93.1 $\pm$ 0.52 & 79.05 & 75.52 $\pm$ 0.59 & 93.9 $\pm$ 0.56\\
& & Single task - Task Perf. & 77.85 & - & - & 79.05 & - & - \\
& & Single task - Concept Perf. & - & 76.11 $\pm$ 0.21 & - & - & 76.07 $\pm$ 0.26 & -\\
& & Single task - Fidelity & - & - & 96.07 $\pm$ 0.49 & - & - & 96.33 $\pm$ 0.26 \\
\midrule
\multirow{8}{*}{\rotatebox[origin=c]{90}{\textbf{Merchant Fraud - NN}}} & \multirow{4}{*}{Ours} & Global & \multirow{4}{*}{74.67} & 82.64 $\pm$ 0.14 & 97.12 $\pm$ 0.29 & \multirow{4}{*}{63.35} & 82.58 $\pm$ 0.12 & 96.62 $\pm$ 0.28 \\
& & Global w/ Ind. &  & 82.6 $\pm$ 0.11 & 96.96 $\pm$ 0.13 & & 82.55 $\pm$ 0.09 & 96.45 $\pm$ 0.24\\
& & Local & & 82.5 $\pm$ 0.14 & 99.39 $\pm$ 0.37 & & 82.45 $\pm$ 0.13 & 99.27 $\pm$ 0.42 \\
& & Local w/ Ind. & & 82.47 $\pm$ 0.13 & 99.34 $\pm$ 0.41 & & 82.42 $\pm$ 0.12 & 99.23 $\pm$ 0.49  \\
\cmidrule{2-9}
& \multirow{4}{*}{Baselines} & Joint CBM ($\lambda$ = 1) & 48.42 $\pm$ 0.31 & 82.49 $\pm$ 0.14 & - & 47.47 $\pm$ 3.64 & 82.34 $\pm$ 0.08 & - \\
& & Distill. Joint CBM ($\lambda$ = 1) & 74.67 & 82.62 $\pm$ 0.13 & 96.87 $\pm$ 0.18 & 63.35 & 82.57 $\pm$ 0.12 & 96.19 $\pm$ 0.29\\
& & Single task - Task Perf. & 74.67 & - & - & 63.35 & - & -\\
& & Single task - Concept Perf. & - & 82.25 $\pm$ 0.19 & - & - & 82.25 $\pm$ 0.19 & - \\
& & Single task - Fidelity & - & - & 98.13 $\pm$ 0.22 & - & - & 97.86 $\pm$ 0.23 \\
\bottomrule
 \end{tabular}
 }
\caption{Mean and standard deviation of results for the CUB-200-2011 (top) and the Merchant Fraud (bottom) NN black-boxes. Results shown for the task performance (Task Perf.), concept accuracy (Concept Perf.) and Fidelity metrics.}
\label{tab:best_dag_results_table}
\end{table}

\begin{figure}[h]
   \subfigure[Learned SCM]{
        \includegraphics[width=0.47\textwidth]{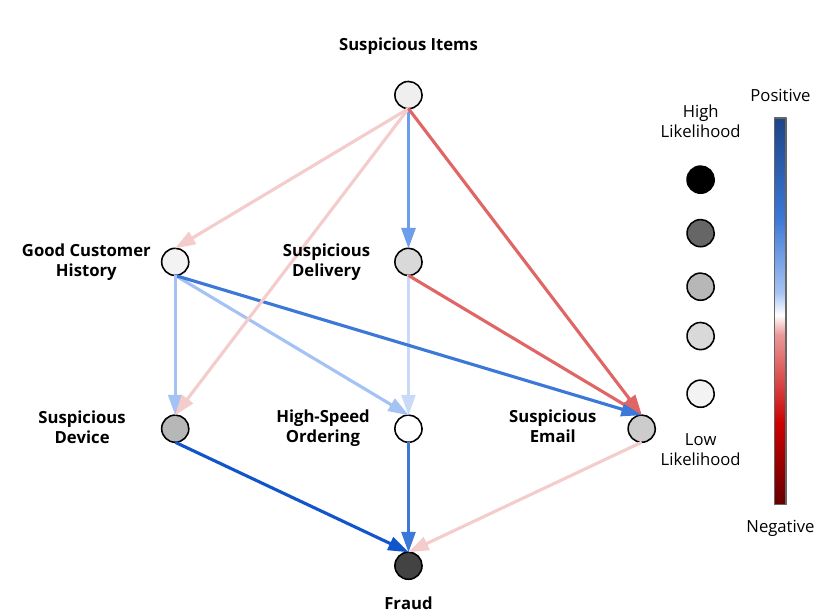}
        \label{fig:structural_explanation}
        }
    \hfill
   \subfigure[Concept attributions]{
        \includegraphics[width=0.47\textwidth]{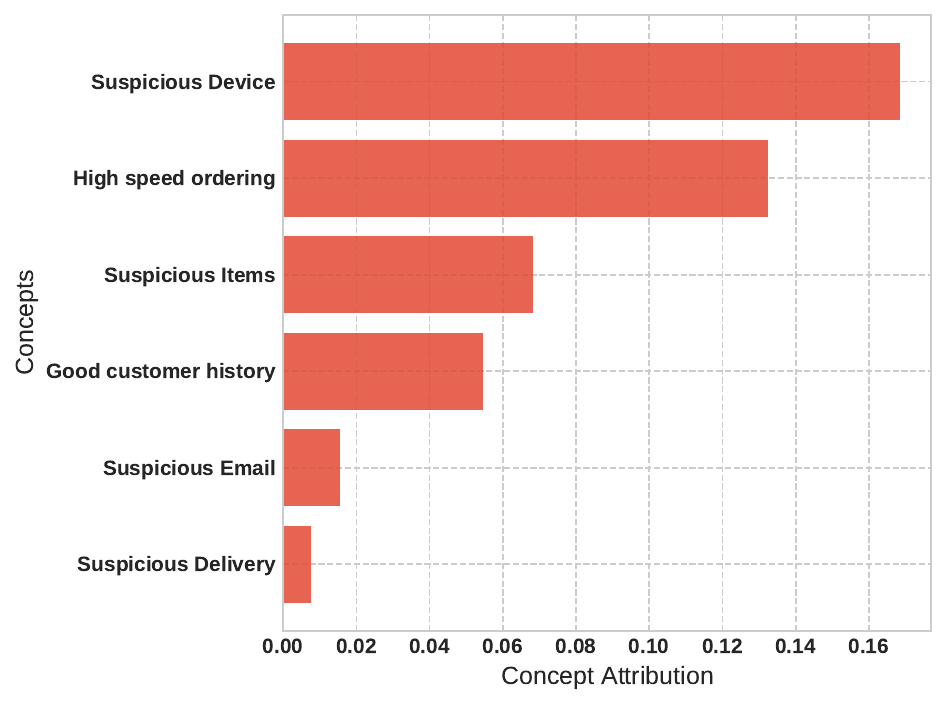}
        \label{fig:attribution_explanation}
        }
    \caption{(a) Learned SCM for an instance of the Merchant Fraud dataset. Blue edge color denotes positive interactions, red denotes negative interactions, and the intensity represents the interaction strength. A positive/negative interaction increases/decreases the value of the destination node, respectively. Concept likelihood (nodes) is encoded from white (low) to black (high). (b)~Concept attribution plot for the same instance.}
     \label{fig:explanation_1}
\end{figure}

\subsection{Local Explanation Example}
Figure \ref{fig:explanation_1} shows an example of a single-instance explanation for the Merchant Fraud dataset in the form of a learned SCM (left) and corresponding concept attributions (right). Concept attributions are obtained by intervening on each concept individually $do(C_k := a), a \in \{0, 1\}$, measuring the total concept effect (TCE) (Equation \ref{eq:tce}) for both interventions and summing their absolute values. Since interventions are performed in the logit space, the assigned values should be in-distribution and associated with the concept being present/not present. Following what is proposed in the CBM paper by \citet{koh2020concept}, we set the values to be the 5th or 95th percentile over the training distribution of each corresponding concept, respectively simulating the absence ($a=0$) or presence ($a=1$) of a chosen concept. We can observe that \textit{Suspicious Device} is the concept that contributes the most for predicting \textit{Fraud} given its strong positive interaction and high concept attribution. Similarly, the absence of \textit{Good Customer History}, a concept we semantically associate with legitimate events, is increasing the likelihood of the children fraudulent concepts. Through these explanations, users can obtain a clearer understanding of not just which concepts are the most relevant for a prediction, but also how they causally affect each other.

\section{Related Work}
Previous works have approached explainability as the task of assessing the causal effects of a model's inputs ($X$) on their outputs ($\hat{Y}$). Most methods focus on the raw input features, providing explanations in the form of feature attributions, but others are closer to our method and focus on concept explanations through causal graphs or by providing concept attributions.

\textbf{Causal feature attribution:} Most methods provide explanations as a set of contributions of each of the model inputs (features). Some works interpret ML models as causal mechanisms, where the input features are the causes and the model predictions are the effects \citep{janzing2020feature, chattopadhyay2019neural, geiger2022inducing, schwab2019cxplain, wang2021shapley, DBLP:journals/corr/abs-2008-00357, frye2020asymmetric, heskes2020causal, jung2022measuring}. With this interpretation,  \citet{frye2020asymmetric, wang2021shapley, jung2022measuring, heskes2020causal} propose extensions of SHAP \citep{lundberg2017unified} that consider the causal structure in the data when calculating the Shapley values. Other methods seek to align neural networks with SCMs, enabling training schemes that produce explainable models \citep{geiger2022inducing} or calculating feature attributions through the use of $do$-calculus \citep{chattopadhyay2019neural}. 
Finally, \citet{schwab2019cxplain} propose a surrogate explainer trained to approximate a black-box model using the Granger-causal objective. Although easily applicable, feature-attribution methods can suffer from a lack of interpretability due to high-dimensionality and lack of knowledge about the feature semantics.

\textbf{Concept explainability:} One of the seminal works in this area is due to \citet{kim2018interpretability}, who developed TCAV. Using TCAV, \citet{ghorbani2019towards} automate the search for concept definitions in images. The CAV method was recently used to mitigate learning of spurious correlations in the data \citep{Wu2023Discover}. Following a different approach, \citet{DBLP:conf/nips/Alvarez-MelisJ18} propose to learn an explainability task together with the classification task, allowing for explanations to be obtained together with predictions at test time. Concept bottleneck models (CBM) \citep{koh2020concept}, work similarly but allow for test-time interventions. Subsequent works have criticized CBMs, claiming the learned concepts lack semantically meaningful input patterns \citep{DBLP:journals/corr/abs-2105-04289}, and that using two objectives results in a trade-off between explainability and performance \citep{DBLP:journals/corr/abs-2205-15480}. The latter work also extends CBMs with a residual predictor that seeks to recover the original main classification performance. Finally, \citet{yeh2020completeness} put forward the definition of \textit{completeness} for assessing how much information the used concepts share with the main classification task, and then propose a methodology to discover completeness-aware concepts. Extending concept-based explanations to tabular data, particularly in fraud detection, \citet{balayan2020teaching} and \citet{belem2021weakly} have proposed a CBM-like approach. This line of work is further expanded by \citet{sousa2022conceptdistil}, who introduced a surrogate explainer approach.

\textbf{Causality for concept explainability:} \citet{DBLP:journals/corr/abs-2006-02482} apply causal discovery to learn a DAG between \textit{concepts} and predictions. \citet{DBLP:conf/iclr/BahadoriH21} propose a causal solution for the lack of \textit{completeness} in CBMs involving adjusting the concept targets as being predicted by the main classification targets. \citet{DBLP:journals/corr/abs-1907-07165} propose a framework that seeks to estimate the causal effects of concepts through interventions; a conditional VAE generates instances where the existing concepts can be intervened on its bottleneck, used to estimate the average causal effect (ACE) of each concept in the black-box model outputs. Similarly, \citet{o2020generative} propose a VAE with a causal influence objective, imposing that some latent space variables influence the outputs of the downstream black-box model, while others are just useful to guarantee a low reconstruction loss. The former variables can be interpreted as concepts learned from the data.

\section{Limitations and future work}

The assumption of a complete set of concepts is strong and it may not be satisfied in a wide set of tasks. In the same vein as what was proposed by \citet{Havasi2022Addressing}, it would be interesting to try future variations of DiConStruct where latent concepts are introduced and allowed to cover any incomplete information. Other possible avenues to relax the need for human labeling are recent methods that exploit language models \citep{Oikarinen2023Label, Yang2023Language}.
Additionally, one limitation of our current setup is that we model each concept and label as a binary. In the future, it would be important to extend the framework to the multi-class case, for example, by using vector-valued nodes in the SCM, changing the sigmoid to softmax activations, and using a multi-class cross-entropy loss instead of the binary cross-entropy.

Recently, it was shown that models with soft concepts can leak information that is detrimental to the interpretability and trustworthiness of the model \citep{Havasi2022Addressing}. Since DiConStruct uses an SCM label predictor, this type of leakage is expected to be somewhat mitigated. Nonetheless, it would be interesting to quantify how much leakage would still be present. Moreover, since we assume a complete set of concepts as well as an SCM capturing the dependencies between concepts, it is argued by \citet{Havasi2022Addressing} that a hard concept version should not compromise task performance while avoiding concept leakage. It would therefore be interesting to compare a hard concept DiConStruct model to the current results.

Since our proposed explainer distills the black-box model, it is prone to copy any learned bias or shortcuts used by that model. Our goal is precisely to create a causal explainer for the black-box predictions: biases should then be surfaced by analyzing the explanations. Different methods could subsequently be applied to mitigate the detected sources of bias.

Recently, \citet{Ghosh2023Dividing, Ghosh2023bridging} proposed MoIE, a post hoc method providing explanations of a black-box model by iteratively extracting interpretable models. These interpretable models in turn provide explanations using first-order logic on the concepts. In future experiments, it would be interesting to compare how the causal explanations from our DiConStruct method relate to the first-order logic relations extracted by MoIE.

\section{Conclusion}
Effective human-AI collaboration can benefit from model explanations expressed through semantic concepts and their causal relations. To this end, we proposed DiConStruct, a surrogate explainer designed to produce explanations that are both causally grounded and concept-based. 

DiConStruct produces explanations in the form of a structural causal model (SCM) describing the interactions between the concepts and the black-box model output. We also obtain concept attributions through counterfactual analysis on the predicted SCM. On two diverse datasets, we show that DiConStruct does not sacrifice performance on the main classification task while simultaneously enriching explainability by facilitating an understanding of the causal relationships between the concepts and the predictions. This is in contrast with state-of-the-art concept explainability methods such as CBM, for which the two objectives of concept prediction and main task performance compete with each other.

\bibliography{causal_diconstruct}

\pagebreak
\appendix
\section{Reproducibility Notes}
\label{reproducibility-notes}
This section provides further details regarding the experiments developed for DiConStruct. Readers can find the code of the experiments of the DiConStruct variants and single task baselines in the folder \texttt{experiments} of the supplementary materials \footnote{\label{footnote-supp-materials-repo} https://anonymous.4open.science/r/causal-diconstruct-4F18/}. The results for causal discovery and the code can also be found in the folder \texttt{experiments}. Finally, the results of the experiments as well as the statistical analysis can be found in the \texttt{results} folder of the supplementary materials \footref{footnote-supp-materials-repo}.

\section{Causality Definitions}
\label{causality-definitions}

Following \citet{pearl2009causality}, a structural causal model (SCM) is defined by a set of $\{x_k\}_{k=1}^K$ endogenous variables, a set of $\{u_k\}_{k=1}^K$ exogeneous variables and a set of deterministic structural functions $\{f_k\}_{k=1}^K$, such that $x_k = f_k(\mathbf{PA}_{k}, u_k)$. Under this formulation, the endogenous variables represent the observable random variables (the concept definitions in our system) and the exogeneous variables represent stochastic unobserved or noise variables. The functions $f_k$ produce structural assignments that depend exclusively on $\mathbf{PA}_k$ (the causal parents of $x_k$) and on the exogeneous variable $u_k$. For causal sufficiency  \citep{pearl2009causality}, the set of exogeneous variables $\{u_k\}_{k=1}^K$ are assumed to be jointly independent, since otherwise there would unobserved variables not captured by the SCM that are causing their dependence, making it unfeasible to perform accurate causal inferences.

A SCM is associated with a directed acyclic graph (DAG) $G = (V,E)$, with the endogeneous variables as the vertices $V = \{x_k\}_{k=1}^K$, and directed edges $E$ denoting direct causation, defined through the functions $f_k$. The causal parents $\mathbf{PA}_k$ of a given endogeneous variable $x_k$ are connected in the graph through a directed edge $\mathbf{PA}_k \rightarrow x_k$.

The operator $do(X := a)$ simulates physical interventions by replacing certain functions from the SCM $\mathfrak{C}$ with constant values, resulting in a new model $\mathfrak{C}_a$, such that $P_{do(X := x)}^{\mathfrak{C}}(Y) = P^{\mathfrak{C}_a}(Y)$. Such interventions are relevant to identify causal effects and perform counterfactual analysis.

\section{A Causal view of Concept explainability}
\label{appendix:causal-view-explainability}
The goal of any explainability system is to provide insights into what information the ML model learned during training. In the case of local explanations, this goal extends to also explaining the decision process followed by the ML model. Similarly to previous works \citep{DBLP:journals/corr/abs-2006-02482, janzing2020feature, chattopadhyay2019neural, wang2021shapley, jung2022measuring, zhao2021causal}, we view that decision process as a complex cause-effect mechanism, $X \rightarrow \hat{Y}$. This interpretation detaches the problem of explainability from the problem of modeling the \textit{real} posterior probability of the label Y given the features X, $P(Y | X)$. We argue that the former is a more well-behaved problem for two reasons: (1) we have a solid causal direction, since we know for a fact that $X$ is a cause for $\hat{Y}$ and (2) $X$ is also the \textit{only} possible cause for $\hat{Y}$, excluding the possibility of unobserved external factors that could act as confounders. 

However, modeling $X \rightarrow \hat{Y}$ as a causal mechanism can still be challenging, considering that $X$ is typically of high dimensionality. Manual or automated causal discovery in such a scenario has a low chance of obtaining a correct causal graph \citep{vowels2022d}. By recentering explainability around concepts $C$ instead of the input features $X$, such a problem is naturally sidestepped, since the dimensionality of $C$ is normally one or two orders of magnitude lower than the dimensionality\footnote{When this is not the case, the humans can always reduce the concept definition (by removing or grouping concepts) to a more digestible number of concepts.} of $X$. Causal discovery in this scenario becomes more tractable, increasing the chance of obtaining a causally correct structural graph. In the same vein of what is proposed in \citep{DBLP:journals/corr/abs-2006-02482}, we model the causal mechanism $C \rightarrow \hat{Y}$ using a causal graph. The possible interactions between the different concepts in $C$ are also described in that causal graph, allowing for the direct and indirect causal effects to be used as concept attributions.

\section{Hyperparameter Details}
\label{hyperparameter-details}

For each of the 48 experiments of the DiConStruct, and also for the concept bottleneck experiments, we performed a random hyperparameter search of 50 iterations. For the remaining models, we perform a separate random hyperparameter search of 200 iterations for each available combination of baseline, black-box model and dataset. For all models, the parameters varied were: (1) number of layers and layer dimensions for all feedforward blocks existing in the architecture\footnote{Common ($L$), concept-specific ($N$), independence discriminator, and local weighting functions.}, (2) dropout probabilities of the dropout layers (3) learning rate, (4) L2 weight decay, (5) use of batch normalization layers, (6) batch size, and (7) categorical feature embedding dimensions (Merchant Fraud dataset only). For the CUB image dataset, we used as the inputs a feature representation of dimension 512 given by the last 8-th layer of a pre-trained ResNet-34 \citep{he2016deep} model.

Additionally, for the experiments with the DiConStruct explainer, we also vary: (8) learning rate of the independence discriminator, (9) the number of epochs to update the independence discriminator, (10) use of the black-box score for the concept distillation SCM, and (11) the existence of biases on the concept distillation SCM

\section{Additional Results}
\label{additional-results}

\subsection{Concept Attribution Diversity Results}
\label{diversity-results}

\begin{figure}[ht]
    \begin{center}
    \includegraphics[width=1.0\linewidth]{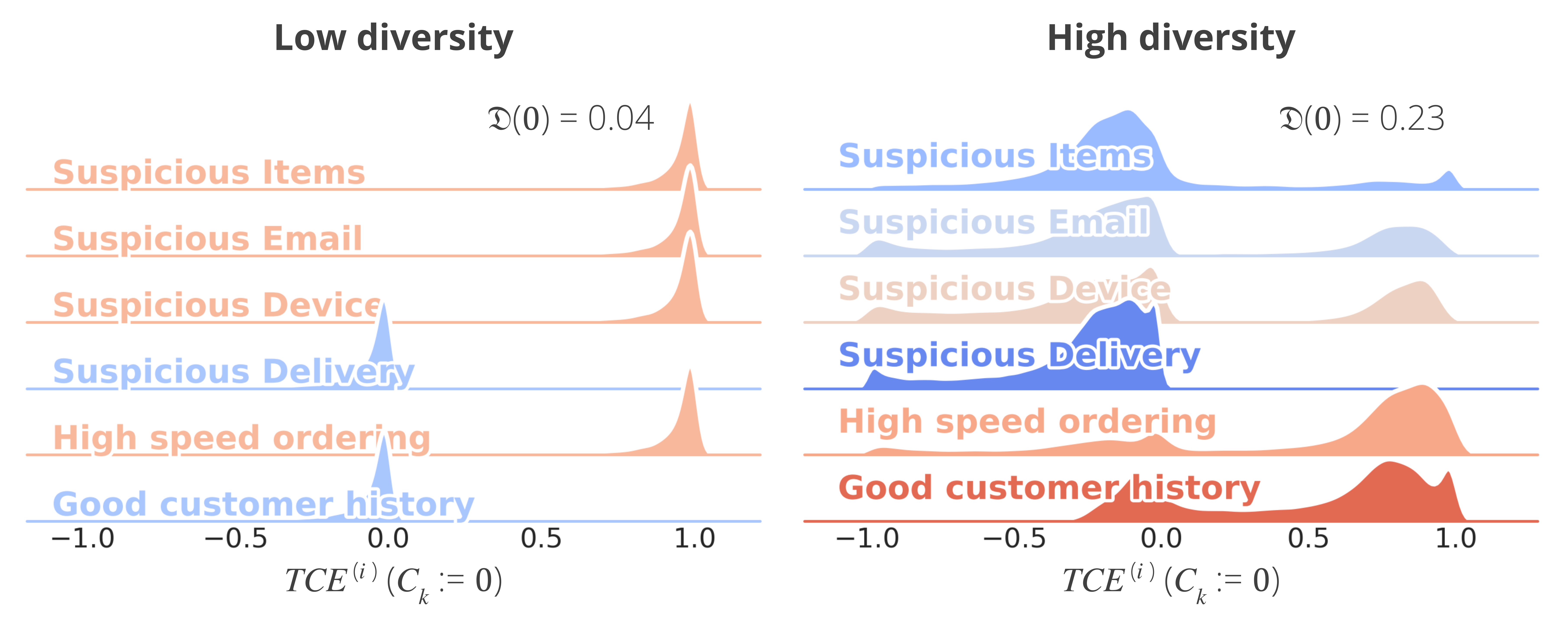}
    \end{center}
    \caption{Concept attribution densities for two DiConStruct explainers trained with the Trivial DAG and evaluated on the Merchant Fraud test set.}
    \label{fig:concept_attribution_diversity}
\end{figure}

\begin{table}[htbp]
\centering
 \caption{Concept attribution diversities for the selection of best results per DAG for the CUB-200-2011 feedforward NN black-box and the Merchant Fraud feedforward NN black-box. Validation and test results are shown for both $\mathfrak{D}(0)$ and $\mathfrak{D}(1)$ diversity metrics. Joint Concept Bottlenecks Models are added for comparison.}
 \resizebox*{\linewidth}{!}{
\begin{tabular}[t]{ll|cccc}
\toprule
\multirow{2}{*}{\textbf{Variant}} & \multirow{2}{*}{\textbf{Best DAG}} & \multicolumn{2}{c}{\textbf{Validation}} & \multicolumn{2}{c}{\textbf{Test}} \\
& & \textbf{$\mathfrak{D}(0)$ (\%)} & \textbf{$\mathfrak{D}(1)$ (\%)} & \textbf{$\mathfrak{D}(0)$ (\%)} & \textbf{$\mathfrak{D}(1)$ (\%)} \\
\midrule
\multicolumn{6}{c}{CUB-200-2011}\\
\midrule 
Global & NO TEARS & 4.94 $\pm$ 0.63 & 4.94 $\pm$ 0.63 & 4.36 $\pm$ 0.61 & 4.36 $\pm$ 0.61 \\
Global w/ Ind. & ICA-LiNGAM & 4.46 $\pm$ 0.64 & 4.45 $\pm$ 0.64 & 3.84 $\pm$ 0.61 & 3.84 $\pm$ 0.61 \\
Local & ICA-LiNGAM & 14.52 $\pm$ 3.79 & 8.77 $\pm$ 3.07 & 12.14 $\pm$ 3.66 & 7.62 $\pm$ 2.97 \\
Local w/ Ind. & PC & \textbf{17.05 $\pm$ 5.38} & \textbf{16.24 $\pm$ 3.37} & \textbf{15.12 $\pm$ 5.19} & \textbf{14.62 $\pm$ 3.42} \\
\midrule
\multicolumn{2}{l|}{Joint CBM ($\lambda$ = 1)} & 4.26 $\pm$ 0.67 & 4.26 $\pm$ 0.67 & 3.74 $\pm$ 0.63 & 3.74 $\pm$ 0.63 \\
\midrule
\multicolumn{6}{c}{Merchant Fraud - NN} \\
\midrule
Global & NO TEARS & 5.77 $\pm$ 0.58 & 5.77 $\pm$ 0.58 & 6.2 $\pm$ 0.64 & 6.2 $\pm$ 0.64 \\
Global w/ Ind. & ICA-LiNGAM & 6.01 $\pm$ 0.26 & 6.01 $\pm$ 0.26 & 6.5 $\pm$ 0.28 & 6.5 $\pm$ 0.28 \\
Local & PC & \textbf{7.89 $\pm$ 2.21} & 30.81 $\pm$ 4.55 & \textbf{8.66 $\pm$ 2.44} & 33.08 $\pm$ 4.88 \\
Local w/ Ind. & PC & 7.02 $\pm$ 1.9 & \textbf{32.06 $\pm$ 5.16} & 7.69 $\pm$ 1.87 & \textbf{34.68 $\pm$ 5.79} \\
\midrule
\multicolumn{2}{l|}{Joint CBM ($\lambda$ = 1)} & 5.91 $\pm$ 0.2 & 5.91 $\pm$ 0.2 & 6.39 $\pm$ 0.32 & 6.39 $\pm$ 0.32 \\
\bottomrule
 \end{tabular}
 }
\label{tab:best_concept_attribution_results_table}
\end{table}

To assess the diversity of the concept attributions given an evaluation set of $M$ instances, $\mathcal{S} = \bigl\{\{\mbox{TCE}^{(i)}(C_k := 0/1)\}_{k=1}^K \in [-1, 1]^K\bigr\}_{i=1}^{M}$, we compute the following metric:
\begin{equation}
    \mathfrak{D}(a) = \frac{1}{KM}\sum_{k=1}^K \sum_{i=1}^M( \mbox{TCE}^{(i)}(C_k := a) - \mu_k^{\mbox{TCE}}(a))^2, \label{eq:diversity}
\end{equation}
where $\mu_k^{\mbox{TCE}}(a)=\frac{1}{M}\sum_{i=1}^M \mbox{TCE}^{(i)}(C_k:=a)$ corresponds to the average TCE of concept $C_k$ over the evaluation dataset.

To understand how different the local concept attribution can be between explainers, we measured both diversity metrics (Equation \ref{eq:diversity}) $\mathfrak{D}(0)$ and $\mathfrak{D}(1)$, on the concept attributions generated for the validation and test datasets. Figure \ref{fig:concept_attribution_diversity} shows the distributions of $TCE^{(i)}(C_k := 0)$ attributions on the test set for two high performing\footnote{We considered explainers with concept accuracy higher than 73\% and $\mathfrak{F}_{MAE}$ higher than 90\%.} explainers of the Merchant Fraud feedforward NN black-box. The explainer on the left side has low concept attribution diversity ($\mathfrak{D}(0) = 0.04$) while the one on the right has high diversity ($\mathfrak{D}(0) = 0.23$). For the explainer with low concept attribution diversity we can observe that, for each individual concept, the attributions are distributed around one central value with low variance, while for the high diversity explainer we have attributions distributed over $[-1, 1]$. 
This result indicates that some explainers may provide very rigid attributions that are close to global explanations. This justifies the need for measuring the concept attribution diversity in our results, we must assess the true locality of the different explainers.

Table \ref{tab:best_concept_attribution_results_table} shows the results for the diversity metrics on the same selection of variants shown in table \ref{tab:best_dag_results_table}\footnote{This selection considers the best explainers in terms of performance, not attribution diversity.}. For computing the local concept attributions $TCE(C_k := 0/1)$ on CBMs, we take the outputs of $g(x)$, intervene on each concept individually ($do(\hat{c}_k := 0/1)$), and calculate the change on the output $f(g(x)_{do(\hat{c}_k := 0/1)}) - f(g(x))$. We can see that the local DiConStruct variant consistently achieves higher concept attribution diversity than the global variants and the CBMs baselines. The complete set of results of each variant and DAG combination is provided in Appendix \ref{complete-performance-results}

\subsection{Complete Set of Results}
\label{complete-performance-results}

\begin{table}[H]
\centering
 \caption{Results for CUB-200-2011 and Merchant Fraud feedforward NN black-box models. The DAGs with the best results, selected for table \ref{tab:best_dag_results_table}, are \underline{underlined}.}
 \resizebox*{\textwidth}{!}{%
\begin{tabular}[t]{ll|cccc}
\toprule
\multirow{2}{*}{\textbf{Variant}} & \multirow{2}{*}{\textbf{DAG}} & \multicolumn{2}{c}{\textbf{Validation}} & \multicolumn{2}{c}{\textbf{Test}} \\
& & \textbf{$\mathfrak{F}_{MAE}$ (\%)} & \textbf{Concept Acc. (\%)} & \textbf{$\mathfrak{F}_{MAE}$ (\%)} & \textbf{Concept Acc. (\%)} \\
\midrule
\multicolumn{6}{c}{CUB-200-2011}\\
\midrule 
\multirow{4}{*}{Global} & ICA-LiNGAM & 92.96 $\pm$ 0.54 & 75.44 $\pm$ 0.46 & 93.88 $\pm$ 0.59 & 75.21 $\pm$ 0.48 \\
& \underline{NO TEARS} & 93.52 $\pm$ 0.77 & 75.58 $\pm$ 0.65 & 94.3 $\pm$ 0.8 & 75.44 $\pm$ 0.44 \\
& PC & 92.1 $\pm$ 0.48 & 75.17 $\pm$ 0.66 & 92.84 $\pm$ 0.51 & 74.99 $\pm$ 0.61 \\
& Trivial & 93.05 $\pm$ 0.51 & 75.61 $\pm$ 0.52 & 93.76 $\pm$ 0.43 & 75.66 $\pm$ 0.48 \\
\midrule 
\multirow{4}{*}{Global w/ Ind.} & \underline{ICA-LiNGAM} & 93.16 $\pm$ 0.75 & 75.61 $\pm$ 0.69 & 93.83 $\pm$ 0.64 & 75.43 $\pm$ 0.65 \\
& NO TEARS & 93.14 $\pm$ 0.57 & 75.21 $\pm$ 0.59 & 93.96 $\pm$ 0.56 & 75.03 $\pm$ 0.73 \\
& PC & 92.45 $\pm$ 0.34 & 75.51 $\pm$ 0.66 & 93.44 $\pm$ 0.34 & 75.42 $\pm$ 0.67 \\
& Trivial & 92.62 $\pm$ 0.46 & 75.91 $\pm$ 0.53 & 93.44 $\pm$ 0.59 & 75.86 $\pm$ 0.57 \\
\midrule 
\multirow{4}{*}{Local} & \underline{ICA-LiNGAM} & 98.72 $\pm$ 0.79 & 75.25 $\pm$ 0.76 & 98.79 $\pm$ 0.74 & 75.11 $\pm$ 0.63 \\
& NO TEARS & 98.65 $\pm$ 0.98 & 74.55 $\pm$ 0.38 & 98.73 $\pm$ 0.92 & 74.22 $\pm$ 0.48 \\
& PC & 98.31 $\pm$ 0.88 & 74.8 $\pm$ 0.85 & 98.37 $\pm$ 0.85 & 74.71 $\pm$ 0.92 \\
& Trivial & 98.53 $\pm$ 0.71 & 75.43 $\pm$ 1.08 & 98.58 $\pm$ 0.7 & 75.33 $\pm$ 1.16 \\
\midrule 
\multirow{4}{*}{Local w/ Ind.} & ICA-LiNGAM & 98.61 $\pm$ 0.73 & 74.69 $\pm$ 0.73 & 98.66 $\pm$ 0.69 & 74.79 $\pm$ 0.89 \\
& NO TEARS & 97.92 $\pm$ 0.99 & 74.27 $\pm$ 0.75 & 97.99 $\pm$ 1.02 & 74.15 $\pm$ 0.92 \\
& \underline{PC} & 98.78 $\pm$ 0.86 & 75.05 $\pm$ 1.08 & 98.83 $\pm$ 0.8 & 74.89 $\pm$ 1.19 \\
& Trivial & 98.4 $\pm$ 0.7 & 75.22 $\pm$ 0.89 & 98.44 $\pm$ 0.7 & 74.91 $\pm$ 0.67 \\
\midrule 
\multicolumn{2}{l|}{Joint CBM ($\lambda$ = 1)} & - & 75.48 $\pm$ 0.53 & - & 75.52 $\pm$ 0.59 \\
\multicolumn{2}{l|}{Single task - Fidelity} & 96.07 $\pm$ 0.49 & - & 96.33 $\pm$ 0.26 & - \\
\multicolumn{2}{l|}{Single task - Concept Perf.} & - & 76.11 $\pm$ 0.21 & - & 76.07 $\pm$ 0.26 \\
\midrule
\multicolumn{6}{c}{Merchant Fraud - NN} \\
\midrule
\multirow{4}{*}{Global} & ICA-LiNGAM & 97.09 $\pm$ 0.22 & 82.58 $\pm$ 0.13 & 96.41 $\pm$ 0.44 & 82.53 $\pm$ 0.11 \\
& \underline{NO TEARS} & 97.12 $\pm$ 0.29 & 82.64 $\pm$ 0.14 & 96.62 $\pm$ 0.28 & 82.58 $\pm$ 0.12 \\
& PC & 96.65 $\pm$ 0.2 & 82.73 $\pm$ 0.17 & 96.06 $\pm$ 0.23 & 82.68 $\pm$ 0.16 \\
& Trivial & 96.91 $\pm$ 0.29 & 82.48 $\pm$ 0.24 & 96.37 $\pm$ 0.27 & 82.43 $\pm$ 0.23 \\
\midrule
\multirow{4}{*}{Global w/ Ind.} & \underline{ICA-LiNGAM} & 96.96 $\pm$ 0.13 & 82.6 $\pm$ 0.11 & 96.45 $\pm$ 0.24 & 82.55 $\pm$ 0.09 \\
& NO TEARS & 96.88 $\pm$ 0.12 & 82.52 $\pm$ 0.11 & 96.4 $\pm$ 0.21 & 82.47 $\pm$ 0.09 \\
& PC & 96.73 $\pm$ 0.27 & 82.6 $\pm$ 0.12 & 95.93 $\pm$ 0.29 & 82.55 $\pm$ 0.09 \\
& Trivial & 96.75 $\pm$ 0.23 & 82.58 $\pm$ 0.13 & 96.25 $\pm$ 0.22 & 82.53 $\pm$ 0.11 \\
\midrule
\multirow{4}{*}{Local} & ICA-LiNGAM & 99.48 $\pm$ 0.38 & 82.27 $\pm$ 0.26 & 99.4 $\pm$ 0.44 & 82.21 $\pm$ 0.25 \\
& NO TEARS & 99.29 $\pm$ 0.41 & 82.47 $\pm$ 0.18 & 99.08 $\pm$ 0.47 & 82.41 $\pm$ 0.17 \\
& \underline{PC} & 99.39 $\pm$ 0.37 & 82.5 $\pm$ 0.14 & 99.27 $\pm$ 0.42 & 82.45 $\pm$ 0.13 \\
& Trivial & 99.48 $\pm$ 0.42 & 82.36 $\pm$ 0.16 & 99.37 $\pm$ 0.48 & 82.31 $\pm$ 0.14 \\
\midrule
\multirow{4}{*}{Local w/ Ind.} & ICA-LiNGAM & 99.34 $\pm$ 0.45 & 82.41 $\pm$ 0.13 & 99.22 $\pm$ 0.51 & 82.36 $\pm$ 0.12 \\
& NO TEARS & 99.33 $\pm$ 0.42 & 82.5 $\pm$ 0.13 & 99.1 $\pm$ 0.52 & 82.45 $\pm$ 0.11 \\
& \underline{PC} & 99.34 $\pm$ 0.41 & 82.47 $\pm$ 0.13 & 99.23 $\pm$ 0.49 & 82.42 $\pm$ 0.12 \\
& Trivial & 99.42 $\pm$ 0.42 & 82.4 $\pm$ 0.17 & 99.27 $\pm$ 0.45 & 82.35 $\pm$ 0.15 \\
\midrule
\multicolumn{2}{l|}{Joint CBM ($\lambda$ = 1)} & - & 82.62 $\pm$ 0.13 & - & 82.57 $\pm$ 0.12 \\
\multicolumn{2}{l|}{Single task - Fidelity} & 98.13 $\pm$ 0.22 & - & 97.86 $\pm$ 0.23 & - \\
\multicolumn{2}{l|}{Single task - Concept Perf.} & - & 82.25 $\pm$ 0.19 & - & 82.25 $\pm$ 0.19 \\
\bottomrule
 \end{tabular}
 }
\label{tab:complete-performance-results}
\end{table}

\begin{table}[H]
\centering
 \caption{Concept attribution diversities for the CUB-200-2011 feedforward NN black-box and the Merchant Fraud feedforward NN black-box. The DAGs with the best results, selected for table \ref{tab:best_dag_results_table}, are \underline{underlined}.}
 \resizebox*{\textwidth}{!}{%
\begin{tabular}[t]{ll|cccc}
\toprule
\multirow{2}{*}{\textbf{Variant}} & \multirow{2}{*}{\textbf{DAG}} & \multicolumn{2}{c}{\textbf{Validation}} & \multicolumn{2}{c}{\textbf{Test}} \\
& & \textbf{$\mathfrak{D}(0)$ (\%)} & \textbf{$\mathfrak{D}(1)$ (\%)} & \textbf{$\mathfrak{D}(0)$ (\%)} & \textbf{$\mathfrak{D}(1)$ (\%)} \\
\midrule
\multicolumn{6}{c}{CUB-200-2011}\\
\midrule 
\multirow{4}{*}{Global} & ICA-LiNGAM & 4.34 $\pm$ 0.57 & 4.34 $\pm$ 0.57 & 3.75 $\pm$ 0.48 & 3.75 $\pm$ 0.48 \\
& \underline{NO TEARS} & 4.94 $\pm$ 0.63 & 4.94 $\pm$ 0.63 & 4.36 $\pm$ 0.61 & 4.36 $\pm$ 0.61 \\
& PC & 4.03 $\pm$ 0.58 & 4.03 $\pm$ 0.58 & 3.55 $\pm$ 0.55 & 3.55 $\pm$ 0.55 \\
& Trivial & 4.41 $\pm$ 0.64 & 4.41 $\pm$ 0.64 & 3.85 $\pm$ 0.59 & 3.85 $\pm$ 0.59 \\
\midrule 
\multirow{4}{*}{Global w/ Ind.} & \underline{ICA-LiNGAM} & 4.46 $\pm$ 0.64 & 4.45 $\pm$ 0.64 & 3.84 $\pm$ 0.61 & 3.84 $\pm$ 0.61 \\
& NO TEARS & 4.97 $\pm$ 0.5 & 4.97 $\pm$ 0.5 & 4.31 $\pm$ 0.43 & 4.31 $\pm$ 0.43 \\
& PC & 4.22 $\pm$ 0.26 & 4.22 $\pm$ 0.26 & 3.73 $\pm$ 0.25 & 3.73 $\pm$ 0.25 \\
& Trivial & 4.55 $\pm$ 0.69 & 4.54 $\pm$ 0.69 & 3.95 $\pm$ 0.66 & 3.94 $\pm$ 0.66 \\
\midrule 
\multirow{4}{*}{Local} & \underline{ICA-LiNGAM} & 14.52 $\pm$ 3.79 & 8.77 $\pm$ 3.07 & 12.14 $\pm$ 3.66 & 7.62 $\pm$ 2.97 \\
& NO TEARS & 15.12 $\pm$ 9.94 & 16.17 $\pm$ 3.97 & 13.5 $\pm$ 8.94 & 14.67 $\pm$ 3.99 \\
& PC & 20.16 $\pm$ 5.97 & 22.27 $\pm$ 5.81 & 18.21 $\pm$ 5.28 & 20.17 $\pm$ 4.96 \\
& Trivial & 15.6 $\pm$ 5.13 & 17.07 $\pm$ 3.0 & 14.06 $\pm$ 5.06 & 15.41 $\pm$ 3.14 \\
\midrule 
\multirow{4}{*}{Local w/ Ind.} & ICA-LiNGAM & 16.74 $\pm$ 4.57 & 10.18 $\pm$ 4.41 & 14.46 $\pm$ 4.38 & 8.96 $\pm$ 4.11 \\
& NO TEARS & 20.88 $\pm$ 11.18 & 18.39 $\pm$ 5.58 & 19.03 $\pm$ 10.65 & 17.5 $\pm$ 6.28 \\
& \underline{PC} & 17.05 $\pm$ 5.38 & 16.24 $\pm$ 3.37 & 15.12 $\pm$ 5.19 & 14.62 $\pm$ 3.42 \\
& Trivial & 19.42 $\pm$ 6.11 & 19.76 $\pm$ 3.41 & 17.85 $\pm$ 5.8 & 18.35 $\pm$ 3.74 \\
\midrule
\multicolumn{2}{l|}{Joint CBM ($\lambda$ = 1)} & 4.26 $\pm$ 0.67 & 4.26 $\pm$ 0.67 & 3.74 $\pm$ 0.63 & 3.74 $\pm$ 0.63 \\
\midrule
\multicolumn{6}{c}{Merchant Fraud - NN} \\
\midrule
\multirow{4}{*}{Global} & ICA-LiNGAM & 6.07 $\pm$ 0.33 & 6.07 $\pm$ 0.34 & 6.48 $\pm$ 0.36 & 6.48 $\pm$ 0.36 \\
& \underline{NO TEARS} & 5.77 $\pm$ 0.58 & 5.77 $\pm$ 0.57 & 6.19 $\pm$ 0.64 & 6.2 $\pm$ 0.64 \\
& PC & 6.01 $\pm$ 0.16 & 6.01 $\pm$ 0.16 & 6.41 $\pm$ 0.21 & 6.41 $\pm$ 0.21 \\
& Trivial & 6.15 $\pm$ 0.2 & 6.15 $\pm$ 0.2 & 6.69 $\pm$ 0.32 & 6.69 $\pm$ 0.32 \\
\midrule
\multirow{4}{*}{Global w/ Ind.} & \underline{ICA-LiNGAM} & 6.01 $\pm$ 0.26 & 6.01 $\pm$ 0.26 & 6.5 $\pm$ 0.28 & 6.5 $\pm$ 0.28 \\
& NO TEARS & 5.78 $\pm$ 0.58 & 5.78 $\pm$ 0.59 & 6.21 $\pm$ 0.62 & 6.2 $\pm$ 0.62 \\
& PC & 6.17 $\pm$ 0.22 & 6.17 $\pm$ 0.22 & 6.65 $\pm$ 0.28 & 6.65 $\pm$ 0.28 \\
& Trivial & 6.08 $\pm$ 0.15 & 6.08 $\pm$ 0.15 & 6.68 $\pm$ 0.18 & 6.68 $\pm$ 0.18 \\
\midrule
\multirow{4}{*}{Local} & ICA-LiNGAM & 9.33 $\pm$ 2.02 & 31.62 $\pm$ 2.75 & 10.3 $\pm$ 2.23 & 33.8 $\pm$ 2.7 \\
& NO TEARS & 12.04 $\pm$ 4.86 & 30.91 $\pm$ 8.49 & 12.87 $\pm$ 5.28 & 32.81 $\pm$ 8.49 \\
& \underline{PC} & 7.89 $\pm$ 2.21 & 30.81 $\pm$ 4.55 & 8.66 $\pm$ 2.44 & 33.08 $\pm$ 4.88 \\
& Trivial & 7.15 $\pm$ 1.59 & 28.38 $\pm$ 5.99 & 7.71 $\pm$ 1.58 & 30.6 $\pm$ 6.2 \\
\midrule
\multirow{4}{*}{Local w/ Ind.} & ICA-LiNGAM & 8.64 $\pm$ 2.45 & 30.38 $\pm$ 3.22 & 9.47 $\pm$ 2.46 & 32.47 $\pm$ 3.75 \\
& NO TEARS & 11.57 $\pm$ 3.09 & 24.95 $\pm$ 6.44 & 13.06 $\pm$ 3.31 & 26.7 $\pm$ 6.85 \\
& \underline{PC} & 7.02 $\pm$ 1.9 & 32.06 $\pm$ 5.16 & 7.69 $\pm$ 1.87 & 34.68 $\pm$ 5.79 \\
& Trivial & 7.67 $\pm$ 1.76 & 31.21 $\pm$ 10.76 & 8.17 $\pm$ 1.87 & 33.51 $\pm$ 11.23 \\
\midrule
\multicolumn{2}{l|}{Joint CBM ($\lambda$ = 1)} & 5.91 $\pm$ 0.2 & 5.91 $\pm$ 0.2 & 6.39 $\pm$ 0.32 & 6.39 $\pm$ 0.32 \\
\bottomrule
 \end{tabular}
 }
\label{tab:complete-diversity-results}
\end{table}

\subsection{Results for Merchant Fraud LightGBM black-box}
\label{lgbm-performance-results}

\begin{figure}[ht]
    \begin{center}
    \includegraphics[width=0.8\linewidth]{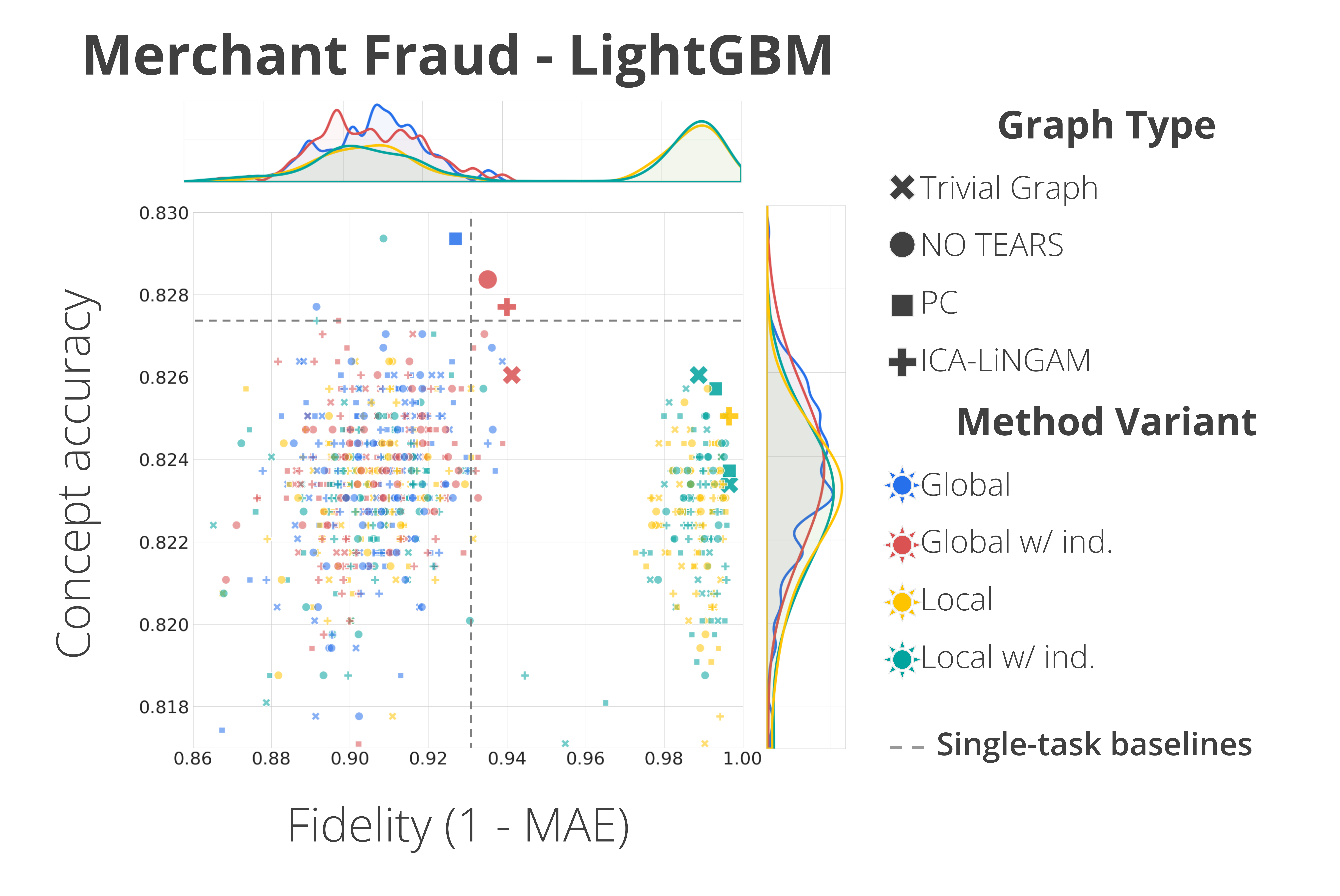}
    \end{center}
    \caption{Distribution of fidelity and concept accuracy for the Merchant Fraud LightGBM hyperparameter experiments. The plot shows the result of 16 experiments of 50 trials, and each marker is a trained DiConStruct explainer. Models on the Pareto efficiency frontier are highlighted with bigger markers.}
    \label{fig:lgbm_performance_scatter_plots}
\end{figure}

The state of the art for classification tasks on tabular data still belongs to the gradient boosted decision tree models \citep{borisov2022deep, shwartztabular, gorishniy2021revisiting, DBLP:conf/iclr/PopovMB20}, like LightGBM \citep{ke2017lightgbm}. For that reason, we included that as a second black-box classifier in our Merchant Fraud dataset experiments. Table \ref{tab:lgbm-results} reports the results of those experiments. The fidelity decreases for Global methods (and single task baseline), when compared to the feedforward NN black-box experiments. We attribute this fidelity decrease in the difference in inductive bias between LightGBM and NN-based models, making the former perform better at the main classification task\footnote{LightGBM performance was 68.31\% recall@5\%FPR against 63.35\% recall@5\%FRP of the fraud NN.}. Intuitively, this means that LightGBM black-box learns more information from $X$ than the NN. In terms of concept prediction performance, we can see that the DiConStruct explainer keeps similar performance to the single task baseline.

Like for the other black-box models, Figure \ref{fig:lgbm_performance_scatter_plots} shows that the DiConStruct explainer outperforms both the knowledge distillation and the explainability single task baselines. For specific configurations, we see that the explainer is better in both tasks than the single task baselines. There is a cluster of local explainers that clearly stands out on the fidelity x-axis. Like with the other black-box experiments, this happens to be a set of explainers configured to use the score of the black-box as input for the weighting functions, $W_{j,k}(\cdot)$, resulting in a much better approximation of the black-box model score.

\begin{table}[htbp]
\centering
 \caption{Main task classification, fidelity and concept prediction performance of DiConStruct explainer for the Merchant Fraud LightGBM black-box. Joint Concept Bottlenecks Models and single task baselines are added for comparison.}
\resizebox*{\textwidth}{!}{%
\begin{tabular}[t]{ll|cccccc}
\toprule
\multirow{2}{*}{\textbf{Variant}} & \multirow{2}{*}{\textbf{DAG}} & \multicolumn{3}{c}{\textbf{Validation}} & \multicolumn{3}{c}{\textbf{Test}} \\
& & \makecell{\textbf{Task} \\ \textbf{Perf. (\%)}} & \makecell{\textbf{Concept} \\ \textbf{Perf. (\%)}} &  $\boldsymbol{\mathfrak{F}_{MAE}} $ \textbf{(\%)} & \makecell{\textbf{Task} \\ \textbf{Perf. (\%)}} & \makecell{\textbf{Concept} \\ \textbf{Perf. (\%)}} & $\boldsymbol{\mathfrak{F}_{MAE}} $ \textbf{(\%)} \\
\midrule
\multicolumn{8}{c}{Merchant Fraud - LightGBM} \\
\midrule
\multirow{4}{*}{Global} & ICA-LiNGAM & 78.36 & 82.6 $\pm$ 0.12 & 93.42 $\pm$ 0.18 & 68.31 & 82.55 $\pm$ 0.1 & 91.98 $\pm$ 1.45\\
& NO TEARS & 78.36 & 82.7 $\pm$ 0.23 & 93.54 $\pm$ 0.18 & 68.31 & 82.65 $\pm$ 0.22 & 92.21 $\pm$ 0.73\\
& PC & 78.36 & 82.65 $\pm$ 0.14 & 93.57 $\pm$ 0.32 & 68.31 & 82.6 $\pm$ 0.12 & 92.48 $\pm$ 1.34 \\
& Trivial & 78.36 & 82.52 $\pm$ 0.11 & 93.62 $\pm$ 0.2 & 68.31 & 82.46 $\pm$ 0.08 & 92.17 $\pm$ 1.17\\
\midrule
\multirow{4}{*}{Global w/ Ind.} & ICA-LiNGAM & 78.36 & 82.57 $\pm$ 0.12 & 93.53 $\pm$ 0.34 & 68.31 & 82.52 $\pm$ 0.1 & 92.56 $\pm$ 1.1 \\
& NO TEARS & 78.36 & 82.56 $\pm$ 0.16 & 93.67 $\pm$ 0.27 & 68.31 & 82.51 $\pm$ 0.14 & 93.01 $\pm$ 0.62 \\
& PC & 78.36 & 82.66 $\pm$ 0.21 & 93.53 $\pm$ 0.17 & 68.31 & 82.61 $\pm$ 0.2 & 92.92 $\pm$ 0.79\\
& Trivial & 78.36 & 82.62 $\pm$ 0.21 & 93.38 $\pm$ 0.33 & 68.31 & 82.57 $\pm$ 0.2 & 91.71 $\pm$ 2.19\\
\midrule
\multirow{4}{*}{Local} & ICA-LiNGAM & 78.36 & 82.43 $\pm$ 0.16 & 99.44 $\pm$ 0.18 & 68.31 & 82.38 $\pm$ 0.15 & 99.42 $\pm$ 0.19\\
& NO TEARS & 78.36 & 82.33 $\pm$ 0.12 & 99.45 $\pm$ 0.18 & 68.31 & 82.28 $\pm$ 0.1 & 99.42 $\pm$ 0.2\\
& PC & 78.36 & 82.4 $\pm$ 0.12 &  99.47 $\pm$ 0.16 & 68.31 & 82.34 $\pm$ 0.1 & 99.4 $\pm$ 0.18 \\
& Trivial & 78.36 & 82.43 $\pm$ 0.13 & 99.48 $\pm$ 0.12 & 68.31 & 82.38 $\pm$ 0.11 & 99.47 $\pm$ 0.15 \\
\midrule
\multirow{4}{*}{Local w/ Ind.} & ICA-LiNGAM & 78.36 & 82.42 $\pm$ 0.13 & 99.41 $\pm$ 0.25 & 68.31 & 82.36 $\pm$ 0.12 & 99.37 $\pm$ 0.26 \\
& NO TEARS & 78.36 & 82.41 $\pm$ 0.17 & 99.48 $\pm$ 0.21 & 68.31 & 82.36 $\pm$ 0.15 & 99.45 $\pm$ 0.22\\
& PC & 78.36 & 82.5 $\pm$ 0.09 & 99.52 $\pm$ 0.15 & 68.31 & 82.45 $\pm$ 0.07 & 99.46 $\pm$ 0.15\\
& Trivial & 78.36 & 82.35 $\pm$ 0.11 & 99.5 $\pm$ 0.17 & 68.31 & 82.3 $\pm$ 0.08 & 99.31 $\pm$ 0.22 \\
\midrule
\multicolumn{2}{l|}{Joint CBM ($\lambda$ = 1)} & 48.42 $\pm$ 0.31 & 82.62 $\pm$ 0.13 & - & 47.47 $\pm$ 3.64 & 82.57 $\pm$ 0.12 & -\\
\multicolumn{2}{l|}{Single task - Fidelity} & - & - & 93.65 $\pm$ 0.31 & - & - & 90.75 $\pm$ 1.19 \\
\multicolumn{2}{l|}{Single task - Task Perf.} & 78.36 & - & - & 68.31 & - & - \\
\multicolumn{2}{l|}{Single task - Concept Perf.} & - & 82.25 $\pm$ 0.19 & - & - & 82.25 $\pm$ 0.19 & -\\
\bottomrule
 \end{tabular}
 }
\label{tab:lgbm-results}
\end{table}

\section{Concept Labels}
\label{sec:concept-labels}

\subsection{CUB-200-2011} 

We used the 28 bird attribute groupings as our concept labels. As described in subsection \ref{datasets-and-blackboxes}, since the original dataset contains multiple possible values per attribute group, they were binarized by assigning 1 to the most common value in the group and 0 otherwise. This resulted in 28 different concepts. To maximize the predictability of our concepts to the black-box score, we selected the top 7 concepts that correlated the most with the black-box model score. We used ROC AUC score between the binary attributes and the continuous score to measure this correlation. The selected list of concepts and corresponding prevalence are shown in table \ref{tab:concept_prevs_cub}.

\begin{table}[htbp]
\centering
 \caption{Concepts for the CUB-200-2011 dataset and corresponding prevalence obtained globally and when conditioning on the classification labels.}
\begin{tabular}[t]{lccc}
\toprule
\multirow{2}{*}{\textbf{Concept}} & \multicolumn{3}{c}{\textbf{Prevalences (\%)}}\\
\cline{2-4}
 & Global  & $Warbler$ & $\neg Warbler$ \\
\midrule
Shape: perching-like  & 48.68 & 76.85 & 44.60 \\
Bill Length: shorter than head & 56.11 & 75.44 & 53.31 \\
Bill Shape: all-purpose & 39.18 & 73.22 & 34.25 \\
Eye Color: black & 83.75 & 91.54 & 82.62 \\
Nape Color: black & 26.48 & 18.86 & 27.58 \\
Primary Color: black & 35.54 & 28.52 & 36.56 \\
Back Color: black & 28.60 & 23.02 & 29.40\\
\bottomrule
\end{tabular}
\label{tab:concept_prevs_cub}
\end{table}

\subsection{Merchant Fraud Dataset} 

To train our method to predict concepts on the full and sampled datasets, we employed a weak supervision methodology. This consisted of creating a set of \textit{Concept Teachers} of the form $f_{CT}: \sX \rightarrow \sY_{C}$ trained using a small golden set of 1934 transactions labeled with concept labels in a procedure described in appendix section \ref{golden-concept-labels}. The \textit{Concept Teachers} were then used to infer the probabilistic labels $p(\vc^{(i)}|\vx^{(i)}) \in [0, 1]$. Our explainability models were trained using these probabilistic labels as concept targets. Training with probabilistic concept targets requires no change of implementation because the BCE loss is equally differentiable for continuous targets $\in [0, 1]$.

\subsection{Merchant Fraud Dataset Golden Concept Labels}
\label{golden-concept-labels}
To evaluate our explainability task, we randomly chose a small set from the sampled dataset, and presented those transactions to a group of human experts (fraud analysts with the knowledge of the most common fraud patterns latent on this online retailer dataset). We then asked the fraud analysts to annotate the existing concepts on the collection of 2643 transactions. These were further split into 1934 for training the \textit{Concept Teachers}, 203 for the optimization of the teachers' parameters, and 506 for test. This last set was the one used to evaluate the explainability performance of the \textit{Concept Teachers} and all DiConStruct models and baselines presented. Despite the requirement for having concept labels being a limitation in this work, we show how this can be mitigated by using a weak supervision strategy. Indeed, we reduced the labelling effort by 3 orders of magnitude, when compared to the effort of fully labelling the sampled dataset of around 500k transactions. Additionally, we consider that concept explainability can benefit from using ground truth concepts obtained from expert knowledge instead of unsupervised methodologies, since the latter can more easily result in meaningless concept definitions.

To provide more context about the concepts considered for the Merchant Fraud, we show in Table \ref{tab:concept_prevs_fraud} the list of concepts as well as their prevalence on the golden dataset (Concept Teachers training data plus the golden test set).

\section{Causal discovery options}
\label{sec:appendix_causal_discovery_options}
\subsection{From Expert Knowledge} 
As a first option, we propose that an iterative work is done together with domain experts. The domain experts should be the same personas involved in concept definition so that there is as familiarity with the problem and an agreement that those are the necessary concepts. Before the procedure starts, a benchmark for the maximum attainable concept prediction accuracy should be created. The procedure is as follows:
\begin{itemize}
    \item An informed guess of $\mathcal{G}_{C,Y_B}$ is drawn by the domain experts, then the DiConStruct explainer is trained using that graph, and its quality is evaluated;
    \item If the distillation fidelity is close to 100\%, the concept prediction accuracy is similar or better than the benchmark and the exogenous factors are jointly independent, the procedure can finish, and the graph structure is saved;
    \item If the distillation fidelity is low, but the concept prediction accuracy is good, the domain experts can try to add more direct parents to $Y_B$, or even remove the causal links $C_k \rightarrow Y_B$ with minimal weights. If after a few iterations on the graph structure, the distillation fidelity keeps low, the problem may be that the concept definition is not \textit{complete}. In such case, the domain experts should assess which new concepts could be added to the concept definition.
    \item In case that both the distillation fidelity and the concept prediction accuracy (compared to the benchmark) are low, the graph structure might be fundamentally incorrect and/or the concept definition is very incomplete. In this scenario, the domain experts can try to redefine $\mathcal{G}_{C,Y_B}$ from scratch or remove the causal links with lower weights and add more direct parents to both the concepts with less prediction performance and to $Y_B$;
    \item If the exogenous independence is not being achieved, and both the distillation fidelity and the concept prediction accuracy are high, it might indicate that some confounding factors are being disregarded in the causal graph. In this scenario, despite the \textit{completeness} and concept prediction accuracy being satisfied, the causal story in incomplete, resulting in less reliable explanations and concept attributions. Therefore, domain experts should assess which new concepts could be added to the concept definition;
\end{itemize}

\subsection{Causal Discovery} 

In case that domain experts are not available, we propose to use a data-driven causal discovery method for obtaining $\mathcal{G}_{C,Y_B}$. This discovery should be constraint structures where $Y_B$ is not causing any of $C$. The dataset used for causal discovery can be the same that will then be used for training the DiConStruct explainer.

\begin{table}[htbp]
\centering
 \caption{Concepts for the Merchant Fraud dataset and corresponding prevalence on the golden dataset. The table shows the prevalence obtained globally and when conditioning on the classification labels.}
\begin{tabular}[t]{lccc}
\toprule
\multirow{2}{*}{\textbf{Concept}} & \multicolumn{3}{c}{\textbf{Prevalences (\%)}}\\
\cline{2-4}
 & Global  & Legit & Fraud \\
\midrule
Good Customer History  & 24.45 & 27.85 & 14.29 \\
High-Speed Ordering & 11.33 & 8.49 & 19.84 \\
Suspicious Delivery & 22.86 & 20.16 & 30.95 \\
Suspicious Device & 11.73 & 8.75 & 20.63 \\
Suspicious Email & 21.07 & 18.93 & 30.16 \\
Suspicious Items & 18.49 & 17.24 & 22.22 \\
\bottomrule
 \end{tabular}
\label{tab:concept_prevs_fraud}
\end{table}

\subsection{Merchant Fraud Concept Teachers} 

To train the \textit{Concept Teachers} we used the golden dataset enriched not only with the fraud features available at run-time but also with some additional information, such as fraud rule triggers and manual decisions, which is only generated during or after the historical fraud decision. The algorithm used for the \textit{Concept Teachers} was a Random Forest classifier and the hyperparameters were optimized using a Tree-structured Parzen Estimator (TPE) \citep{bergstra2011algorithms} algorithm with 200 trials from which 30 trails were used for a random initialization. As the optimization metric, we used ROC AUC on the golden validation set. We then evaluated the \textit{Concept Teachers} in the test set of our golden dataset, obtaining a mean ROC AUC of 75.72\% across all concepts. Despite not being central to our work, the \textit{Concept Teachers} performance shows an interesting result when compared to the 78.49\% performance of our concept explainabilty baseline. This 2.8 pp improvement originates solely from the using more data in $\sX$ (the concept explainability baseline trained in 460k instances labeled by our \textit{Concept Teachers}) which illustrates that our weak supervision approach is successful to some degree.

\end{document}